# MILD: Mediator Agentic System with Bidirectional Perception and Multi-Layered Alignment for Human-Vehicle Collaboration

Jiyao Wang[1], Yunbiao Wang[1], Yubo Jiao[1], Xiao Yang[2], Dengbo He[2], Sasan Jafarnejad[3], Luis Miranda-Moreno[1], Raphaël Frank[3] , and Jiangbo Yu[1,✉]

**ABSTRACT:** Prior studies report that partial driving automation can increase the cognitive demands on human drivers. This effect largely arises from human drivers' lack of transparent insight into the vehicle's intentions and decision logic, as well as from automated systems' limited awareness of the driver's dynamic state and preferences. This bidirectional misalignment undermines shared situational awareness and exacerbates coordination failures in human-vehicle interaction. To address these limitations, we argue for a paradigm shift that elevates the human role from passive supervisor to active manager. We introduce the Mediator-in-the-Loop-Driving (MILD) system, based on an agentic system architecture to facilitate synergistic human-vehicle collaboration. MILD integrates a perception agent for joint in-cabin and out-of-cabin understanding with a lightweight strategy agent that generates compliant and explainable action suggestions. To ensure these strategies are strictly aligned with safety regulations and human values, we develop Evidence- and Constraint-weighted Policy Optimization (ECPO). ECPO leverages automatic validators to steer the agent toward behaviors that are not only accurate but also structurally complete, substantiated by evidence, and less prone to constraint violations. Furthermore, a retrieval-augmented generation module dynamically incorporates constraints from traffic regulations, speed recommendations, and driver preferences into the decision loop. Experiments across three open datasets demonstrate that MILD consistently outperforms baselines in both perception accuracy and strategy quality under auditable offline metrics, and yields higher human-rated policy adequacy, comfort, and explanation than baselines. These results support MILD as a practical reference design for auditable high-level intelligent cockpit mediation and offer insights for building future human-vehicle collaborative driving.

**KEYWORDS:** human-vehicle collaboration, large language models, intelligent cockpit, agentic system, policy optimization, multi-layered alignment

## 1 Introduction

The evolution from manual to automated driving represents a fundamental paradigm shift in the human-vehicle relationship (Bengler et al., 2014; Fagnant and Kockelman, 2015). As codified by the SAE levels of automation (SAE On-Road Automated Vehicle Standards Committee, 2018) and shown in **Fig. 1**, this transition has progressively recast the human's role from a direct controller of the vehicle to a supervisor of an increasingly autonomous system (Badue et al., 2021). While this advancement has offloaded the physical task of driving, it has introduced new cognitive challenges (Stapel et al., 2019; Ange Wang et al., 2025). The supervisory role is often ill-defined, placing significant demands on the human to monitor both the vehicle's performance and the external environment, ready to intervene when necessary. A primary limitation of this paradigm is the weak bidirectionality of communication: the human supervises the vehicle, but the vehicle has limited access to the human's fluctuating internal state, creating a critical collaboration gap within the human-advanced driving system (ADS) team (Hasenjäger et al., 2020).

To transcend the limitations of the direct supervision model, we argue that a new paradigm is necessary: one that elevates the human's role from a supervisor to a manager. In this paradigm, the human would not only rely on automation for low-level vehicle control, as is increasingly the case in SAE L2 systems, but also delegate the configuration of assistance functions, human-machine interface (HMI) channels, and comfort-related actuators to an intelligent agent that can negotiate the distribution of tasks and constraints within the human-vehicle team. This conceptual shift is grounded in the well-documented shortcomings of traditional supervisory control in dynamic domains, where a passive human monitor can suffer from out-of-the-loop performance problems, leading to degraded situational awareness and delayed response times in critical moments (Endsley and Kiris, 1995). For instance, an SAE L2 vehicle operating in its preset "assertive" mode in dense traffic may induce significant anxiety in the human driver; blind to this cognitive state, the vehicle continues its aggressive maneuvers, increasing the risk of an erratic human takeover or simply eroding trust. This highlights a need to move towards more collaborative frameworks, which advocate for a continuous and dynamic negotiation between the human and the machine (Marcano et al., 2020; Parasuraman and Riley, 1997).

Most current research implicitly assumes this asymmetric and weakly bidirectional communication pattern. Studies for driving personalization aimed to make the supervised vehicle more attuned to its user (e.g., personalized adaptive cruise control (Wang et al., 2022) or lane-change assistance (Zhu et al., 2018). However, these systems typically learn static driving "styles" from past behavior and fail to adapt to the dynamic, intra-trip cognitive and emotional states of the supervising human (Bao and Li, 2025). This creates a misalignment that can undermine comfort and erode trust in the automation (Stilgoe and Mladenović, 2022). On the other hand, driver monitoring systems (DMS) attempt to reverse the supervisory channel, allowing the vehicle to monitor the human (Wang et al., 2024; Zhang et al., 2022), but they are usually isolated alarm mechanisms that trigger standardized alerts without deeply integrating external traffic context or vehicle capabilities (Haas and van Erp, 2014; Xu et al., 2025). Consequently, current systems can increasingly "see" the road and, to some extent, the driver, but lack a principled way to translate this joint understanding into auditable, constraint-aware strategies. Recent advances in Large Language Models (LLMs) in transportation (Qu et al., 2023) and human sensing (Lu et al., 2024) have motivated first attempts at such mediating agents. However, their approach relies on relatively passive and limited in-cabin information sensing and does not explicitly model multi-level constraints or joint in-/out-of-cabin perception, and auditing and evaluating such systems from human factors and regulatory perspectives remains challenging.

[1] Department of Civil Engineering, McGill University, Montreal H3A 0G4, Quebec, Canada [2] Systems Hub, The Hong Kong University of Science and Technology (Guangzhou), Guangzhou 511453, Guangdong, China [3] Interdisciplinary Centre for Security, Reliability and Trust, University of Luxembourg L-4365, Luxembourg

✉ Corresponding author. E-mail: jiangbo.yu@mcgill.ca

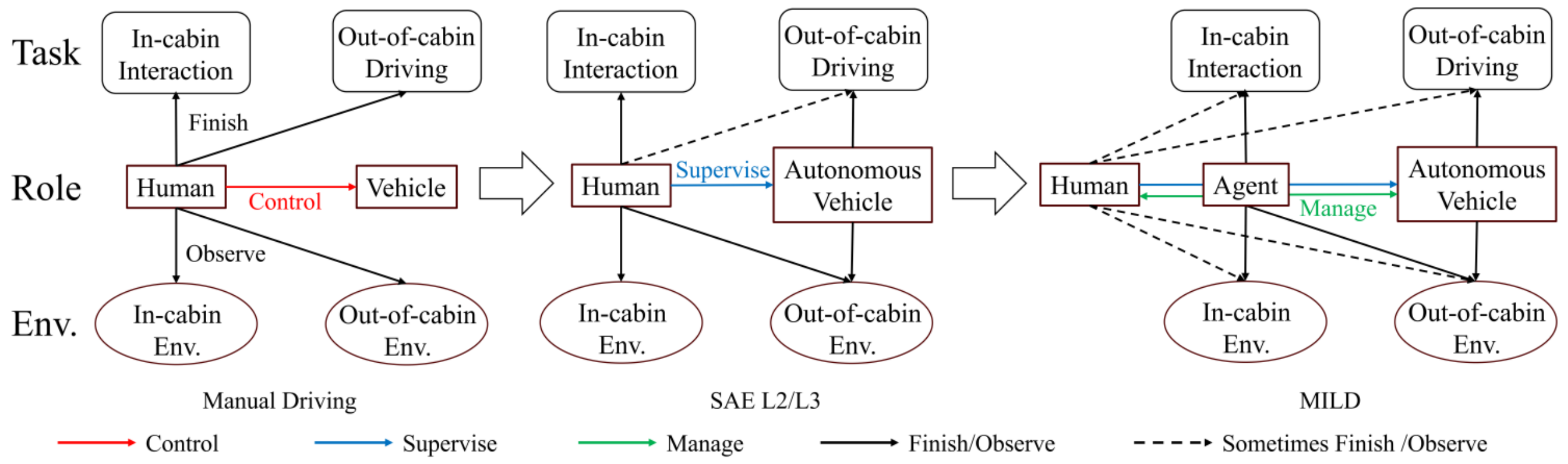

**Fig. 1.** Conceptual evolution of the human–vehicle relationship from (a) direct manual control, where the driver operates the vehicle without mediation, to (b) supervisory use of an ADS, where the human monitors an opaque automation stack, and finally to (c) the proposed Mediator-in-the-Loop-Driving (MILD) paradigm, in which an LLM-based agent mediates between the driver and the ADS.

In this work, we introduce the Mediator-in-the-Loop- Driving (MILD) system, an agentic vehicular system with a perception agent and a strategy agent. The perception agent is responsible for low-level perception tasks, specifically jointly understanding in- cabin and out-of-cabin environments. It integrates classical perception tasks with scene understanding, creating a unified output consisting of discrete labels and structured sample summaries that provide a consistent interface for downstream policy learning. The strategy agent serves as the central decision-making component of the system. It is a lightweight LLM that offers compliant, explainable, and traceable action suggestions to the driver and ADS by engaging with humans, environments, and external tools or resources. These suggestions are reasoned based on the agent's understanding of the vehicle's current internal and external environment, the available in-cabin tools, and retrieved resources such as legal regulations, vehicle limits and driver preferences.

To move beyond black-box decision making and simple accuracy optimization, we further propose Evidence-and Constraint-weighted Policy Optimization (ECPO), a validator-induced weighted pairwise preference objective inspired by DPO. Instead of relying on a separate learned reward model or an explicit frozen reference policy, ECPO derives preferences directly from auditable validators and is optimized jointly with supervised fine-tuning, so that constraint compliance, evidence grounding, and structural completeness become first-class training signals. A retrieval-augmented generation (RAG) module injects up-to-date, multi-level constraints drawn from local traffic regulations, vehicle specifications and limits, and driver preferences, while environment-related conditions are grounded in the joint in-/out-of-cabin perception. We evaluate this design with a unified offline protocol across three open datasets, reporting both ECPO-derived policy metrics and traditional perception benchmarks. Results demonstrate that MILD surpasses strong baselines in estimation accuracy and in alignment with safety- and human-centric objectives. In summary, our work offers the following contributions:

- We introduce MILD, an agentic system in the vehicle that realizes bidirectional understanding and strategic support between driver and vehicle, built on a unified joint in-/out-of-cabin perception target and a lightweight strategy agent.
- We operationalize multi-layered alignment for driving agents via a constraint-grounded policy language, retrieval-based constraints, and auditable validators, enabling unified offline evaluation of schema validity, constraint compliance, and evidence grounding, complemented by human ratings.
- We instantiate MILD with ECPO, an evidence- and constraint-weighted preference objective, to further strengthen the strategy agent's alignment capacity while ensuring auditable processes.
- We conduct extensive experiments across three datasets. The results show that MILD improves perception accuracy and strategy quality under auditable metrics, and that MILD achieves better human-centric objective evaluations than baselines.

## 2 Related Works

### *2.1. Large Language Models for Autonomous Driving*

Multi-modal LLMs are reshaping ADS by shifting from optimization-centric modular pipelines to knowledge-driven systems with stronger reasoning capability (Cui, Ma, et al., 2024). Compared with purely data-driven methods that often fail in long-tail scenarios, LLM-based agents can leverage broad world knowledge to interpret complex traffic contexts and synthesize more human-like driving policies.

Recent studies apply LLMs to motion planning and vehicle control. GPT-Driver (Mao et al., 2023) casts motion planning as language modeling by tokenizing heterogeneous sensory inputs and directly predicting ego trajectories. DiLu (Wen et al., 2024) augments this paradigm with a memory module that stores and reflects on driving experiences, improving closed-loop generalization. To bridge high-level reasoning and low-level feasibility, LanguageMPC (Sha et al., 2023) uses an LLM to adapt MPC parameter matrices, maintaining kinematic consistency. DriveGPT4 (Xu et al., 2024) further advances interpretable end-to-end driving by fusing multi-view video and textual queries to predict low-level control signals, showing improved performance in dynamic scenes. DriVLMe (Huang et al., 2024) explores navigation via embodied experience and social dialogue, enabling route planning from natural-language instructions.

In parallel, benchmarks and datasets emphasize reasoning and explainability. DriveLM (Sima et al., 2024) formulates driving as a perception-prediction-planning reasoning chain within a graph-based VQA benchmark to answer "why" questions about decisions. Reason2Drive (Nie et al., 2024) provides over 600K video-text pairs to evaluate chain-of-thought reasoning with perceptual grounding in complex environments. Extending to 3D, OmniDrive (Shihao Wang et al., 2025) employs counterfactual reasoning to anticipate hazards and generate 3D-aware explanations. For trajectory prediction, LC-LLM (Peng et al., 2025) uses supervised fine-tuning to infer lane-change intentions with textual justifications, improving transparency in interaction-aware models.

Despite these advances, existing methods remain largely vehicle-centric, often producing technically safe yet user- misaligned behaviors due to limited modeling of driver internal states, and a unified bidirectional framework that jointly reasons over traffic context and driver state is still absent.

### *2.2. Human-Centric Intelligent Cockpits and Agents*

As autonomous driving transitions from vehicle-centric control to human-centered services, Intelligent Cockpit research has progressed from simple command execution toward robust driver-state understanding and proactive interaction (Li et al., 2024). Current efforts largely fall into three intertwined directions: multimodal driver monitoring, personalized interaction agents, and LLM-enabled cognitive assistance.

In driver state monitoring, recent studies infer drivers' affective states and cognitive workload from complementary modalities For instance, DrowsyDG-Video combines multi-view facial landmark augmentation with adversarial domain generalization for robust driver drowsiness detection across subjects and recording setups (Li et al., 2026). MD-EPN integrates textual, acoustic, and visual cues for multi-dimensional emotion recognition (Xie and Xie, 2024), while Wang et al. (2024) estimates driver states together with heart and respiration rates directly from RGB facial video, reducing dependence on wearable sensors. Building on this line, attention-based audio-video fusion improves robustness under complex driving conditions where single-modality cues are unreliable (Luan et al., 2025). Beyond camera-based approaches, DrowsyDG-Phys employs domain generalization over electrocardiogram, electrodermal activity, and respiration signals to achieve robust cross-domain drowsiness detection (Jiyao Wang, Li, et al., 2026), and A. Wang et al., (2026) fuse the same three physiological modalities via an aligned-attention transformer to estimate cognitive load in conditionally automated vehicles.

In personalized interaction and agents, LLMs (often coupled with vision-language models) enable cockpit assistants to resolve underspecified instructions and provide user-tailored feedback. Representative frameworks map natural-language requests to vehicle actions while leveraging long-term memory to learn individual driving preferences such as headway and acceleration style (Cui, Yang, et al., 2024); persistent memory combined with dynamic user profiling further supports continual adaptation across extended interactions (Westhäußer et al., 2025). LLM-PDA adopts retrieval-augmented generation to personalize multimodal warning policies across visual, auditory, and haptic channels (Xu et al., 2025). In contrast, PACE-ADS proposes a multi-agent collaboration design and modulates vehicle behavior by inferring user mental states from physiological signals (Bao and Li, 2025), while Sage Deer explores multi-view reasoning for micro-expressions and body movements directly over multi-view video data (Lu et al., 2025).

Despite progress, current approaches are limited by reliance on intrusive sensing and by the lack of a unified schema that semantically links external traffic events with in-cabin driver responses for causal, user-aware decision support.

### *2.3. Policy Alignment and Constrained Generation*

Ensuring that LLM-generated driving policies comply with safety regulations and human values is essential in autonomous driving, yet standard next-token prediction can yield hallucinated or unsafe outputs (Yue Zhang et al., 2025). Alignment methods therefore optimize models toward safer behaviors using human preference data, most notably RLHF (Dai et al., 2023), while DPO has recently gained traction as a more stable and compute-efficient alternative that removes the need for an explicit reward model by directly fitting preferences (Rafailov et al., 2023).

In autonomous driving, these ideas have been adapted to domain-specific safety objectives: PE-RLHF injects physics-informed feedback by incorporating safety constraints into the reward design (Huang et al., 2025), and SafeAuto further reduces violations by embedding explicit safety knowledge (e.g., traffic rules) into multimodal LLMs via a reasoning-enhanced auxiliary task (Jiawei Zhang et al., 2025). Complementary to preference-based alignment, constrained generation enforces structural or semantic requirements at inference time, ranging from satisfying large-scale logical constraints (Boffa and You, 2025) to adhering to strict output schemas for tool use (Qin et al., 2023)-and, in driving control, LanguageMPC constrains LLM decisions by adapting MPC parameters to remain within kinematic feasibility (Sha et al., 2023).

However, existing alignment treats safety as a soft, non-hierarchical objective and remains largely non-auditable, motivating validator-driven frameworks that enforce hard constraints and yield verifiable compliance.

## 3 Methodology

In this section, we present the technical details of the proposed MILD

system. The overall framework of MILD is shown in **Fig. 2**. We first formalize the background setting and problem formulation. We then describe how the perception agent unifies in-/out-of-cabin understanding into a joint target, how the schema-driven strategy agent turns this target into auditable strategies, and how the ECPO objective aligns the strategy agent with multi-level constraints. Finally, we introduce the retrieval-augmented constraint module and implementation details. For clarity, **Table 1** summarizes the key symbols used throughout this paper.

### *3.1 Background and Problem Formulation*

We consider the current mainstream semi-automated driving settings (i.e., SAE L2), where humans and automation jointly bear responsibility for safety, comfort, and progress. The vehicle is equipped with in-cabin sensors (e.g., multi-view driver-facing cameras) and out-of-cabin sensors (e.g., surround-view cameras), as well as an assistance stack providing longitudinal/lateral control and various comfort and HMI actuators. To avoid introducing new safety or liability issues, the agent in this work does not issue low-level control commands. Instead, it produces high-level, structured suggestions for (i) driver-facing communication (e.g., warnings, explanations), (ii) driving suggestion (e.g., driving mode selection, car-following gap) when an assistance stack is present, and (iii) comfort and HMI actuators such as heating, ventilation, and air conditioning (HVAC), audio, or ambient light.

Conceptually, we view the closed-loop driving process as a partially observed sequential decision problem. The underlying environment state lives in a state space $\mathcal{S}$, while the agent only has access to an observation space $\mathcal{O} = \mathcal{O}^{in} \times \mathcal{O}^{out} \times \mathcal{P}^{drv} \times \mathcal{P}^{veh} \times \mathcal{K}$, which combines in-cabin $\mathcal{O}^{in}$ and out-of-cabin observations $\mathcal{O}^{out}$, driver $\mathcal{P}^{drv}$ and vehicle profiles $\mathcal{P}^{veh}$, and applicable constraint snippets $\mathcal{K}$. The high-level action space is formed by composing a small set of atomic action types $\mathcal{A}$ (e.g., driving advice, driver communication, HVAC, audio, ambient light) with type-specific parameter spaces, temporal horizons, and priorities. These actions are ultimately executed by the underlying tool space $\mathcal{T}$, which groups concrete actuators such as the assistance stack, HMI channels, and comfort actuators. MILD operates in this abstracted observation--action interface: it observes $\mathcal{O}$, selects high-level actions in $\mathcal{A}$, and relies on $\mathcal{T}$ to apply them, without issuing low-level control commands.

Formally, let $x^{in} \in \mathcal{O}^{in}$ and $x^{out} \in \mathcal{O}^{out}$ denote the multi-view in-/out-of-cabin observations over a temporal window, $p^{drv} \in \mathcal{P}^{drv}$ and $p^{veh} \in \mathcal{P}^{veh}$ denote driver and vehicle profiles (e.g., risk tolerance, experience, available assistance functions), and $k \in \mathcal{K}$ denote text snippets describing applicable constraints (e.g., traffic regulations, vehicle limitations and specifications, driver preferences). A mediator-in-the-loop sample is represented as:

$$e = (x^{in}, x^{out}, p^{drv}, p^{veh}, k) \tag{1}$$

Given a sample $e$, the goal of MILD is to produce a structured high-level strategy $y^{pol}$ in a predefined policy language.

$$y^{pol} = \pi_\theta(\mathrm{e}), y^{pol} \in \mathcal{Y}^{\mathrm{pol}} \tag{2}$$

Here $\mathcal{Y}^{pol}$ is a constrained structured space of multi-step strategies constructed from atomic actions in $\mathcal{A}$. The human driver retains authority over the vehicle: $y^{pol}$ is surfaced through HMI and assistance configuration in $\mathcal{T}$ rather than direct throttle, brake, or steering commands.

**Table 1. Summary of key symbols used in MILD.**

| Symbol | Description |
|---|---|
| | Spaces and sets |
| $S$ | Environment state space. |
| $O$ | Observation space seen by MILD. |
| $P$ | Profile space (driver or vehicle.) |
| $K$ | Space of textual constraint snippets (laws, vehicle specifications and manuals, driver preferences). |
| $A$ | Set of atomic high-level action types (ADS advice, HMI, HVAC, etc.). |
| $T$ | Tool/actuator space (assistance stack, HMI channels, comfort actuators). |
| $Y^{pol}$ | Space of structured PolicyAction schema objects. |
| $D$ | Training dataset $\{(e_i, y_i^{ref})\}_{i=1}^{N}$. |
| N | Number of samples in $D$ |
| | Variables and representations |
| $x$ | Observations over a temporal window. |
| $p$ | Profile for a given sample. |
| $k$ | Retrieved and compressed constraint summary. |
| $e$ | Mediator-in-the-loop sample $e=(x^{in}, x^{out}, p^{drv}, p^{veh}, k)$ |
| $z$ | Joint perception representation |
| $\ell$ | Discrete labels for perception. |
| $s^{joint}$ | Three-stage joint summary (initial, transition, final). |
| $u$ | Strategy input bundle $(z, p^{drv}, p^{veh}, k)$. |
| | Models |
| $g_\phi$ | Perception agent with parameters $\phi$. |
| $\pi_\theta$ | Strategy agent with parameters $\theta$. |
| | Policy and actions |
| $y^{pol}$ | Policy instance produced by $\pi_\theta$ in $Y^{pol}$. |
| $y^{ref}$ | Reference policy from the cleaned labels. |
| $a_j$ | $j$-th atomic action in a policy (type, parameters, etc.). |
| $J$ | Number of actions in a policy. |
| $y^{(k)}$ | $k$-th candidate policy sampled. |
| $y^+, y^-$ | Best and worst policies for a given prompt $u$. |
| | Losses and optimization |
| $\mathcal{L}_{percep}$ | Perception SFT loss for $g_\phi$. |
| $\mathcal{L}_{SFT}$ | Strategy SFT loss for $\pi_\theta$. |
| $\mathcal{L}_{ECPO}$ | ECPO-based preference optimization loss. |
| $\mathcal{L}_{strategy}$ | Final strategy training objective. |
| $\lambda_{ECPO}$ | Weight of ECPO loss in $\mathrm{L}_{strategy}$. |
| $f_\theta(u,y)$ | Log-unnormalized score of $y$ under $\pi_\theta$. |
| $w(u)$ | ECPO-based weight for preference pair at prompt $u$. |
| $\psi(\cdot)$ | Monotone mapping from ECPO gap to weight $w(u)$. |
| | Validator and ECPO metrics |
| $L(y)$ | Violation severity level of policy $y$(0-4). |
| $C(y)$ | Violation count of policy $y$. |
| $s_{core}(u,y)$ | Core safety score derived from $L(y)$ and $C(y)$. |
| $s_{evd}(u,y)$ | Evidence coverage score for $y$ given $u$. |
| $s_{str}(u,y)$ | Structural completeness score for $y$. |
| ECPO$(u,y)$ | Aggregated ECPO score in [0,1]. |

We learn the strategy policy $\pi_\theta$ from an offline dataset $\mathcal{D} = \{(e_i, y_i^{ref})\}_{i=1}^{N}$, where each $y_i^{ref} \in \mathcal{Y}^{pol}$ is a teacher-assisted reference strategy. Concretely, an initial policy draft is produced by GPT-5[1] and then manually audited and revised into a cleaned target that satisfies the PolicyAction schema, respects the retrieved constraints, and refers back to perceptual evidence. In our data construction process, 68.5% of generated policies required at least one manual revision, indicating that teacher outputs were not directly accepted as final labels. We therefore use this teacher-assisted construction as a scalable supervision pipeline for feasible and auditable high-level strategies, rather than as an unquestioned source of ground truth or a unique optimal policy.

[1] https://openai.com/gpt-5/

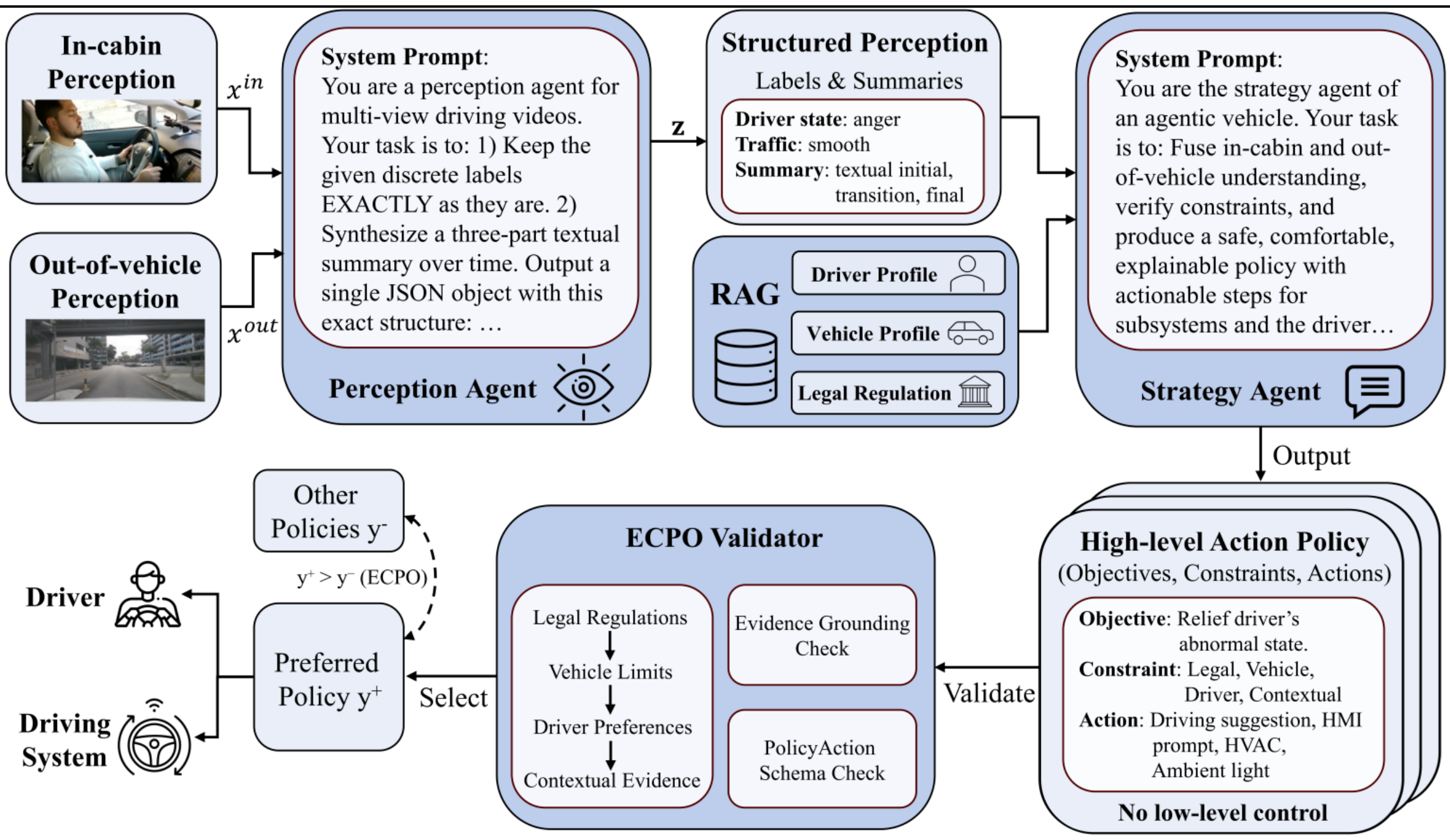

**Fig. 2**. Overview of MILD framework. At runtime, a perception agent processes in-cabin and external sensory inputs to produce structured outputs with discrete labels and temporal summaries, which are fused with retrieved driver and vehicle profiles and legal regulations. A strategy agent then generates high-level driving policies following a strict schema of objectives, constraints, and actions, explicitly excluding low-level vehicle control. During training, candidate policies are evaluated by an ECPO validator, which enforces multi-source constraints and derives preference signals to select the preferred policy.

### *3.2 Overview of the MILD Architecture*

The MILD architecture follows a perception-strategy-alignment pattern. At a high level, it decomposes the Mediator-in-the-Loop-Driving problem into three coupled components:

- a **perception agent** that converts raw in-/out-of-cabin observations into a compact joint representation $z$;
- a **strategy agent** that maps $z$, profiles and constraints to structured high-level strategies $y^{pol}$ in the policy language;
- a **constraint and alignment layer** that couples retrieval-augmented constraints with the ECPO objective to enforce multi-level compliance.

This design is inspired by recent LLM-based driver agents that separate perception, control, and action modules for personalized warnings (Xu et al., 2025), but extends them in two directions. First, we ask the perception agent to jointly summarize in- and out-of-cabin states in a single target tailored to policy learning. Second, we treat constraint satisfaction and evidence grounding as first-class optimization signals rather than post-hoc checks.

Concretely, the perception agent $g_\phi$ receives the sensor streams and produces a joint structured object $z$ that combines discrete labels and structured sample summaries:

$$z = g_\phi(x^{in}, x^{out}). \tag{3}$$

The strategy agent then operates on a textual bundle and outputs a policy $y^{pol} = \pi_\theta(u)$ under the policy language.

$$u = (z, p^{drv}, p^{veh}, k), \quad y^{pol} = \pi_\theta(u). \tag{4}$$

### *3.3 Joint In-/Out-of-cabin Perception Agent*

#### 3.3.1 Unified perception target

Instead of attaching and fusing multiple task-specific heads downstream, MILD unifies perception already at the representation level. For each sample, the perception agent produces $z = (\ell^{drv}, \ell^{scene}, s^{joint})$, where $\ell^{drv}$ are driver-related labels (e.g., distraction, wheel grip, gaze allocation), $\ell^{scene}$ are scene-related labels (e.g., ego maneuver, traffic density), and $s^{joint}$ is a three-stage textual summary with fields: Initial, Transition, and Final state. All of these are encoded in a constrained structured template with fixed keys and value formats.

This design has three consequences. First, it forces the agent to always report both in-cabin and out-of-cabin states, avoiding the tendency of specialized models to ignore ``the other side" of the vehicle. Second, it eliminates optional fields and free-form formatting that are difficult for a downstream policy learner to exploit. Third, by constraining the vocabulary and structure of z, it reduces the risk that training data will leak dataset-specific fingerprints into model outputs, which is important for privacy and licensing.

#### 3.3.2 Training the perception agent

Following recent research in autonomous driving and driver monitoring (Cui, Ma, et al., 2024) and considering the limited computational resources in vehicles, we implement $g_\phi$ as a small (7–8 billion parameters) multimodal large language model (MLLM) (details are illustrated in Section 4.1. For in-cabin data, the base model learns to predict multi-task driver monitoring labels and short rationales; for out-of-cabin data, it learns to describe dynamic objects and scene evolution; for manual driving samples, it learns to rewrite human summaries into the unified three-stage structure.

All three datasets are converted into the same target format. Given input $x = (x^{in}, x^{out})$ and the corresponding ground-truth perception output $z^\star$, we fine-tune $g_\phi$ with a causal language modelling objective $\mathcal{L}_{\text{percep}}(\phi)$ with supervision restricted to tokens inside the structured object.

$$\mathcal{L}_{\text{percep}}(\phi) = -\mathbb{E}_{(x,z^\star)}\left[\log p_\phi(z^\star \mid x)\right]. \tag{5}$$

In practice, we adopt parameter-efficient adaptation, using Low-Rank Adaptation (LoRA) (Hu et al., 2021) for supervised fine-tuning (SFT), which freezes the pre-trained model weights and injects trainable low-rank matrices into each Transformer layer.

### *3.4 Schema-driven Strategy Agent*

#### 3.4.1 Constraint-grounded policy language

To make strategies auditable, executable by vehicle-side tools, and comparable under a unified offline protocol, we define a schema-driven policy language, $PolicyAction$ schema (**Fig. 3**), instead of free-form text. It is the complete field-level contract used for (i) training targets, (ii) runtime parsing, and (iii) all ECPO/validator checks reported. A policy instance is:

$$y^{pol} = (objectives, constraints, actions). \tag{6}$$

where $objectives$ summarizes the intended safety/human-centric goal in one compact statement, $constraints$ logs the constraint layers that were explicitly consulted, and $actions$ is a short sequence ($1 \le J \le J_{\max}$) of atomic high-level actions, where $J_{\max}$ the maximum number of allowed actions per policy.

```
{
// Section 1: Objectives & Constraints
"objectives": { //A concise objective statement grounded in the current situation.
      "e.g., Address reduced visibility and maintain safe time headway..."},
"constraints": [
      "Legal Regulations": <Traffic rule or regulation that must not be violated>",
      "Vehicle Limits": <Capability / actuator limit that has been verified>",
      "Driver Preferences": <Preference or sensitivity from driver profile>",
      "Contextual Evidence": <Contextual or scenario-specific constraint (e.g., weather, road type)>"],

// Section 2: Action Sequence
"actions": [
   {
     "type": "Driving suggest| HMI prompt | HVAC | AmbientLight",
     "parameters": {
           "modality": "visual | audio | multimodal",
           "text": "Visibility reduced by rain. Please maintain a larger distance...",
           // Examples for HVAC / AmbientLight etc.:
                 // "target temperature": 23,
                 // "fan level": "low",
                 // "ambient theme": "cool focus", ...},

     "rationale": "e.g., Explain why this action is chosen given driver state, scene, and constraints.",

     "evidence":
        "in cabin text": [
            "e.g., 'driver shows signs of anxiety and frequent head turns"],
        "out of vehicle text": [
            "e.g., 'heavy rain with limited visibility and dense traffic ahead"],
        "objects": [
            "Referenced object IDs and descriptors from out-of-vehicle perception"],
        "labels": [
            e.g., 'emotion: anxious', 'behavior: looking around']}
}
```

**Fig. 3.**The illustration of PolicyAction schema.

Each action $a_j$ follows the fixed keys in **Fig. 3**: $\{type, parameters, rationale, evidence\}$. The $type$ field selects a high-level channel (Driving suggestion/HMI/HVAC /Ambient light), while "$parameters$" holds type-specific settings. $rationale$ must explain why the action is chosen given driver state, scene context, and constraints, and $evidence$ must reference concrete elements from$z$ and/or retrieved snippets $k$ (e.g., in-cabin text, out-of-cabin text, object IDs, and labels). We disallow impermissible low-level control at the language boundary: any explicit throttle/brake/steering command (or equivalent control token) invalidates the output.

### *3.5 Evidence- and Constraint-weighted Policy Optimization (ECPO)*

We introduce ECPO, an evidence- and constraint-weighted preference regularization objective inspired by DPO (Rafailov et al., 2023), but tailored to multi-level constraints and validator-derived preferences. Unlike classical DPO, ECPO is not used here as a standalone replacement for supervised learning. Instead, it serves as an auxiliary alignment term on top of strategy supervised fine-tuning: the strategy agent is first anchored to cleaned reference policies, while ECPO further biases the model toward candidates that better satisfy layered constraints, evidence grounding, and structural completeness. Rather than learning a separate reward model, we directly compare candidate policies using validators and use these comparisons to regularize the strategy policy. Throughout this subsection, $y$ always denotes a policy instance in $\mathcal{Y}^{pol}$.

**Table 2.** Validator inventory and violation severity mapping. The validator checks layered constraints in descending priority and assigns $L(y) \in \{0, \ldots, 4\}$ based on the highest-priority violated layer.

| Layer | Source | Example checks | $L(y)$ |
|---|---|---|---|
| Legal regulations | Local traffic codes in $k$ | stop/yield compliance; prohibited maneuvers; HMI legal limits | 4 |
| Vehicle limits | Vehicle profile + manuals in $k$ | unavailable actuator; mode capability mismatch; rate limits | 3 |
| Driver preferences | Driver profile in $p^{drv}$+$k$ | discomfort triggers; preferred HMI channel; motion-sickness constraints | 2 |
| Contextual evidence | Joint perception $z$ | hazard-aware conservatism; maneuver-context consistency | 1 |
| None | – | no violations | 0 |

#### 3.5.1 Validator and ECPO score

Evaluating strategy agents in driving is fundamentally different from benchmarking perception models. Even for human–centric criteria such as workload, comfort, or trust, there is rarely a single ground-truth action sequence or universally agreed quantitative metric. Recent LLM-based mediators for in-cabin assistance (Bao and Li, 2025; Xu et al., 2025) largely rely on qualitative user studies, small-scale expert ratings, or case-based analyses, which are difficult to reproduce and do not directly provide a dense training signal for policy learning. Our goal is therefore to build a relatively stable, quantitatively defined evaluation layer that is still grounded in domain knowledge and human factors, and that can be used both for offline assessment and for shaping the strategy agent.

Given a prompt $u$ and a candidate policy $y$, a non-differentiable validation suite scores $y$ along four axes. The validator first checks $y$ for alignment with four-layered constraints, in order of priority: Legal regulations, Vehicle limits, Driver preferences, and Contextual evidence. The first three layers are instantiated through the retrieved snippets $k \in \mathcal{K}$, which summarize traffic laws, vehicle specifications and limits, and driver-specific preferences. The environment layer is instantiated through the perception outputs $z$.

The validator returns a discrete violation severity level $L(y) \in \{0, \ldots, 4\}$, while $C(y)$counts the number of distinct violated checks across all layers. **Table 2** summarizes the validator inventory. The severity level $L(y)$ is determined by the highest-priority violated layer (legal $\succ$ vehicle $\succ$ driver $\succ$ contextual), while $C(y)$ counts distinct violated checks across all layers. Specifically, legal regulations define a hard safety boundary that cannot be overridden. Vehicle limits describe what the system can physically and functionally achieve inside that legal envelope. Contextual-related constraints ensure that, the policy is explicitly grounded in the provided context (structured perception and retrieved constraints) rather than being speculative. The ordering (legal ≻ vehicle ≻ driver ≻ contextual) follows decreasing non-negotiability and risk externality: legal and vehicle constraints are priority requirements with direct safety and feasibility implications, driver constraints aim to satisfy driver preferences within the admissible envelope, and contextual-evidence grounding primarily improves traceability under potentially incomplete or noisy context. Importantly, the four layers are not enforced as strict hard constraints in the sense of immediately terminating evaluation once a violation appears. Instead, the validator audits the policy sequentially in descending priority and applies hierarchical penalties: violations at earlier, higher-priority layers lead to a larger severity level, while additional violated checks further reduce the overall core score through $C(y)$. Thus, ECPO does not treat the layered constraints as binary pass/fail gates, but as a priority-sensitive penalty structure. These outputs are merged into a core safety score $s_{core}(u, y)$.

$$s_{core}(u, y) = \max\left(0, 1 - \frac{L(y)}{4} - 0.1 \min(C(y), 10)\right). \quad (7)$$

Multiple violations are thus penalized almost as strongly as a single severe violation, and $s_{core}(u, y) \in [0,1]$ with higher values indicating fewer or milder violations under the layered validator.

An evidence coverage module then computes $s_{evd}(u, y) \in [0,1]$ by checking, for each action, how many of the entities and clauses listed in its evidence field can be matched back to the joint perception summaries $z$ and the retrieved constraint snippets $k$, and averaging this fraction across actions. A structural completeness score $s_{str}(u, y) \in [0,1]$ measure whether $y$ is valid under the policy schema, whether the action list is non-empty and not excessively long, and whether each action is accompanied by a rationale. We start from 1 and subtract fixed penalties for invalid or missing fields. These components are then combined into a normalized $\mathrm{ECPO}(u, y)$.

$$\mathrm{ECPO}(u, y) = 0.5 s_{core}(u, y) + 0.3 s_{evd}(u, y) + 0.2 s_{str}(u, y). \quad (8)$$

Constraint compliance remains the dominant signal, while evidence grounding and structural completeness act as auxiliary regularizers that improve auditability and policy usability. We adopt $(s_{core}, s_{evd}, s_{str}) = (0.5, 0.3, 0.2)$ as a balanced default and further verify through sensitivity analysis that the qualitative behavior of ECPO is stable under nearby settings. The score $\mathrm{ECPO}(u, y)$ lies in [0,1] by construction. Higher scores correspond to fewer or milder violations, stronger evidence grounding and better structural quality. ECPO encodes this hierarchy by assigning higher violation levels to breaches of higher-priority layers in $L(y)$, and by rewarding policies that explicitly ground each action in evidence from both $z$ and $k$ through $s_{evd}$ and $s_{str}$.

### 3.5.2 ECPO-based preference optimization

Even with the validator and the ECPO score in place, a natural question is why we need a preference-based optimization stage in addition to supervised fine-tuning. Pure SFT trains the strategy agent to mimic given ground-truth policies $y^{ref}$, but any systematic biases or occasional constraint violations in the current closed-source LLM are inevitable. Moreover, SFT gives each training example roughly equal weight, regardless of how strongly it satisfies or violates the four-layer constraints, and it cannot directly incorporate the non-differentiable validator into the learning signal. Classical DPO (Rafailov et al., 2023) addresses some of these issues by learning from preference pairs, yet it typically assumes externally provided human preferences or a learned reward model, and it does not explicitly encode layered safety and human-centric constraints. Our ECPO-based preference optimization extends this line of work by deriving preferences directly from the validator and by weighting pairs according to the ECPO gap, so that prompts with a clear safety and alignment signal have a stronger influence on the policy.

Given a strategy prompt $u = (z, p^{drv}, p^{veh}, k)$, we sample $K$ candidate policies $\{y^{(i)}\}^{K}$, where each $y^{(i)}$ is a policy instance generated under the schema. Each candidate is scored by the ECPO validator, yielding an aggregated score $\mathrm{ECPO}(u, y^{(i)})$. We denote by $y^{+}$ and $y^{-}$ the highest-scoring and lowest-scoring candidates for the same prompt, respectively.

$$y^{+} = \arg\max_{i} \mathrm{ECPO}(u, y^{(i)}), \quad y^{-} = \arg\min_{i} \mathrm{ECPO}(u, y^{(i)}). \quad (9)$$

This procedure constructs preference pairs $(y^{+}, y^{-})$ without human comparison labels. The preferences are induced entirely by domain-specific, non-differentiable validators that encode driving safety, constraint compliance, and evidence grounding. In contrast to standard DPO (Rafailov et al., 2023), which assumes externally provided preference pairs (typically from human raters or a learned reward model), ECPO turns a multi-objective constraint checker into an automatic preference generator tailored to in-cabin agents. We train the strategy agent with a preference objective, driven by these ECPO-derived pairs.

$$\mathcal{L}_{ECPO} = -\mathbb{E}_{u,(y^{+},y^{-})}\left[w(u) \log \sigma\left(\beta\left(f_{\theta}(u, y^{+}) - f_{\theta}(u, y^{-})\right)\right)\right] \quad (10)$$

In this equation, $f_{\theta}(u, y)$ denotes the log-unnormalized score that $\pi_{\theta}$ assigns to y given $u$, $\sigma(\cdot)$ is the logistic function, and $\beta > 0$ is a temperature controlling how sharply the model is pushed to separate preferred from dispreferred policies. The scalar weight $w(u)$ depends on the ECPO gap for that prompt:

$$w(u) = \psi(\mathrm{ECPO}(u, y^{+}) - \mathrm{ECPO}(u, y^{-})). \quad (11)$$

The mapping $\psi(\cdot)$ is monotonically increasing (for example, a clipped linear mapping). Samples where the validator suite strongly prefers one policy over another thus contribute more to the gradient update.

Finally, we combine imitation and ECPO-based preference optimization into the overall training objective for the strategy agent.

$$\mathcal{L}_{strategy} = \mathcal{L}_{SFT} + \lambda_{ECPO} \mathcal{L}_{ECPO}. \quad (12)$$

In this formulation, ECPO functions as a constraint-aware alignment regularizer rather than an independent policy-learning objective. The main anchor of the strategy policy remains the cleaned reference strategy $y^{ref}$ used in supervised fine-tuning, while $\mathcal{L}_{ECPO}$ adds extra pressure toward validator-preferred candidates when multiple plausible schema-valid policies exist. Unlike classical DPO, in our setting, the policy space is already strongly constrained by the PolicyAction schema, the training data provide cleaned reference strategies, and the ECPO term is only used to regularize the model toward safer and better-grounded outputs. The hyperparameter $\lambda_{ECPO}$ controls how strongly ECPO regularization influences the policy relative to supervised teacher matching.

ECPO training and runtime validator play complementary roles rather than replacing one another. During training, ECPO is used to reshape the model's output distribution, making structured, valid, human-aligned, and better-grounded policies easier to generate. At deployment time, the ECPO validator can be applied as a post-hoc rule-based safety layer to further audit or gate generated policies. In this sense, ECPO improves the policy model upstream, while runtime validation remains the downstream guarantee.

### *3.6 Retrieval-augmented Constraint Injection*

Traffic regulations, vehicle capability limits, and driver preference guidelines are dynamic (market-dependent and frequently updated). Encoding them purely in model parameters would make maintenance brittle and expensive. Following the externalized constraint-memory design used in prior personalized driver-agent work (Xu et al., 2025), MILD maintains a lightweight, versioned constraint store and injects only the relevant constraints into the decision loop via retrieval-augmented generation (RAG) (Lewis et al., 2020).

The RAG contains short and structured texts grouped into three sources: (i) jurisdiction-specific traffic rules, (ii) vehicle specifications/manual-derived limits and actuator availability, and (iii) driver preference/profile guidelines. Each passage is stored with provenance metadata (e.g., source category, jurisdiction/market, vehicle configuration, and an internal clause ID), enabling policies to log which constraints were consulted. Specifically, given $u = (z, p^{drv}, p^{veh}, k)$, we form a retrieval query using jurisdiction and operating mode (from $p^{veh}$), driver sensitivities/preferences (from $p^{drv}$), and situation cues/hazards summarized in z. For retrieval, we use a frozen off-the-shelf dense bi-encoder retriever with cosine similarity. The retriever selects the most relevant snippets $k$, which are then condensed into a compact constraint summary that fits within the strategy context. The strategy agent is required to (i) record the checked layers and referenced clauses in `constraints` and (ii) cite supporting evidence in `evidence`. This makes the decision process auditable: retrieved rules shape generation, and ECPO validators explicitly check compliance against the same retrieved constraints.

## 4 Experiments

### *4.1. Implementation Details*

We implement MILD using parameter-efficient fine-tuning on recent multi-modal and text-only LLMs. For the perception agent $g_\phi$, we select **Qwen3 − VL − 8B** (Bai et al., 2025), **LLaVA − NeXT − Video − 7B** (Li et al., 2024), and **InternVL − 3.5 − 8B** (Weiyun Wang et al., 2025) for their balance of video understanding capability and automotive-grade deployment feasibility. For the strategy agent $\pi_\theta$, we employ compact reasoning models: **Qwen3 − 4B** (Yang et al., 2025) and **Llama − 3.2 − 3B** (Grattafiori et al., 2024). All baselines, including GPT-5, receive identical schema definitions, profile/constraint fields, and evidence citation instructions to ensure fair comparison. The RAG module is implemented as a frozen external retriever of MILD. We use BAAI/bge-base-en-v1.5 as the dense bi-encoder retriever.

For training, we freeze backbone weights and inject LoRA modules (rank=32, $\alpha$=16) into attention and feed-forward layers. Training is conducted via LLaMa Factory (Zheng et al., 2024) on 2 × NVIDIA A800 GPUs using Bfloat16 precision, AdamW optimizer (lr=$5e-5$, cosine schedule), and a 5k token context window. The pipeline consists of two stages: (1) **Perception**, where $g_\phi$ is fine-tuned to map raw inputs to the unified target z (loss masked to schema tokens only); and (2) **StrategyLearning**, where $\pi_\theta$ is fine-tuned on policy samples via $\mathscr{L}_{SFT}$ followed by $\mathscr{L}_{ECPO}$. Particularly, for the strategy agent ECPO training, the candidate policy number K is set to 2. Inference of agents uses temperature=0.7, top-p=0.8, and max new tokens=5096. At runtime, the perception agent operates on overlapping temporal clips rather than frame-wise inputs. Each refreshed clip is converted into one joint structured summary, which then triggers one policy refresh by the strategy agent. In our implementation, the perception agent uses a 10 s sliding window with a 2 s step. This means that the policy layer is refreshed at the summary-update rate, rather than at the low-level vehicle-control frequency. The clip duration is chosen to be long enough to support the Initial–Transition–Final representation while remaining shorter than the timescale of high-level assistance adaptation.

### *4.2 Datasets*

We utilize three datasets converted into our unified perception target z and policy input u:

*DMD* is a large-scale, naturalistic driver monitoring dataset that uses multiple cameras and wearable sensors to record drivers over forty hours of real-world driving, involving over thirty participants (Ortega et al., 2020). In our pipeline, we retain the in-cabin camera streams, remap category labels to a multi-label taxonomy, and generate three-part textual summaries (Initial, Transition, and Final state) from the videos together with the dataset-provided discrete labels, followed by manual review for consistency and grounding. This delivers richly supervised in-cabin data, but does not include external scene information.

*DriveLM* is based on large-scale datasets such as nuScenes (Caesar et al., 2020), provides a language-based benchmark for autonomous driving. It represents driving scenes as graphs and supplies QA pairs about perception, prediction, planning, and driver behavior from a bird's-eye view (Sima et al., 2024). We construct out-of-cabin samples using multi-camera footage and scene annotations. For each temporal window, we derive object lists and a three-stage textual summary in our Initial–Transition–Final format from the graph-based question–answer annotations of DriveLM. The summary jointly describes scene elements, predicted trajectories, maneuvers, and the behavior of nearby traffic, and populates $\ell^{scene}$ and the scene-related parts of $s^{joint}$; in-cabin footage is not available.

*AIDE* is a recent dataset for advanced driver assistance with synchronized in-cabin and outward-facing cameras (Yang et al., 2023). It spans urban and highway scenarios, normal and safety-critical events, and provides discrete labels for driver state, vehicle maneuvers, and the traffic scene. We convert each sample into joint driver and scene labels, along with a three-stage summary, as with DMD. These summaries are constructed by verbalizing synchronized videos together with the dataset's discrete driver/scene annotations into a unified textual format, followed by manual review.

Synthetic pairing of DMD and DriveLM (Mixed). To stress-test whether a single perception–strategy–alignment pattern can generalize across heterogeneous corpora, we construct a synthetic paired dataset (Mixed) by combining in-cabin samples from DMD and out-of-cabin samples from DriveLM. Specifically, if $n_{dmd}$ and $n_{dlm}$ are the numbers of samples, each DMD sample is parsed into structured driver labels and a three-stage summary, and each DriveLM sample into object lists and scene summaries in the same Initial–Transition–Final format. To reduce deterministic pairing artifacts, we first shuffle DriveLM samples within each split (train/val/test) and then pair each DMD sample with one DriveLM sample (or a small block) using a split-preserving mapping. We repeat the pairing with multiple random seeds and vary the block size in a small range; all reported Mixed results are averaged over these pairings to mitigate spurious correlations by single pairing rule. Nevertheless, the following results on Mixed do not imply deployable joint reasoning, but only distribution-robustness under schema-aligned supervision.

Each synthetic pair is assigned driver and vehicle profiles, sampled evenly, and augmented with legal rules as well as available actuators, which are transformed through the RAG module into constraint text k. This yields a synthetic in-/out-of-cabin dataset with joint context for policy learning and evaluation, used primarily for robustness checks and controlled comparisons rather than as a substitute for truly paired data.

### *4.3 Evaluation Metrics*

*Perception Metrics*. In DMD and DriveLM, several perception tasks are formulated as example-based multi-label classification problems. We evaluate the perception agent's performance with example-based Intersection over Union (IoU), exact match ratio (EMR), and sample-wise F1 score (F1) (all in \%). Specifically, on DMD, we treat the three driver-state groups (distraction, hands-on-wheel, gaze) as multi-label sets. On DriveLM, object recognition is cast as a set prediction problem over identified traffic participants. For a dataset with $N$ samples, let $Y_i$ denote the ground-truth label set (or object set) for sample $i$, and $\hat{Y}_i$is the corresponding prediction. The sample-wise F1 score is computed for each sample and then averaged across the dataset. When the label set and prediction are both empty, the sample is excluded from IoU/EMR/F1 aggregation. The definitions of these metrics are, where $\epsilon$ is a small constant to avoid division by zero:

$$IoU = \frac{1}{N}\sum_{i=1}^{N}\frac{|Y_i \cap \hat{Y}_i|}{|Y_i \cup \hat{Y}_i|}, \quad (13)$$

$$EMR = \frac{1}{N}\sum_{i=1}^{N}\mathbf{1}\,[Y_i = \hat{Y}_i], \quad (14)$$

$$F1 = \frac{1}{N}\sum_{i=1}^{N}\frac{2|Y_i \cap \hat{Y}_i|}{|Y_i| + |\hat{Y}_i| + \epsilon}. \quad (15)$$

The four AIDE classification tasks (driver emotion, driver behavior, traffic scene, vehicle motion) are treated as standard single-label classification problems. For each task, every sample $i$ has a ground-truth class $y_i \in \{1, \dots, C\}$ and a predicted class $\hat{y}_i$, where $C$ is the number of classes. The accuracy is:

$$\mathrm{Acc} = \frac{1}{N}\sum_{i=1}^{N}\mathbf{1}\,[y_i = \hat{y}_i]. \quad (16)$$

For AIDE, we report Accuracy (Acc) and macro F1 for each of the four classification tasks, both in percentage form (i.e., multiplied by 100).

The perception agent also produces structured textual summaries. On DMD and AIDE, these take the form of a three-stage narrative with fields: Initial, Transition, and Final state. On DriveLM, we map all textual question-answer pairs into the same three-stage format, and then evaluate each stage with the same metrics. For each textual field, we compare predicted and reference summaries using three corpus-level metrics: BLEU-4 (Papineni et al., 2002), ROUGE-L (Lin, 2004), and BertScore (Zhang* et al., 2019).

*Strategy evaluation*. Intuitively, the strategy metrics answer five complementary questions: Can the model produce a parseable policy at all (Valid percentage **Valid**%)? Does it violate higher-priority constraints (Violation severity **ViolSev**)? Does it overstep into forbidden low-level vehicle control (Low-level control rate **LowCtrl**%)? Does it explicitly address the key hazards present in the current context (Hazard evidence coverage F1 **HazF1**)? And do independent human auditors judge the policy as compliant, safe, and evidence-grounded (Human Audit Score **HAS**)? Valid%, ViolSev, LowCtrl%, and HazF1 are offline, auditable proxies derived from schema checks, layered constraint validation, and hazard set matching, while HAS provides an evaluation signal that is independent of the ECPO training-time validator.

**Valid**% measures basic output reliability. Given a driving situation prompt $u$ and a candidate policy $y$, we first check whether $y$ is a syntactically valid instance of the $PolicyAction$ schema. Valid% is the fraction of test prompts for which this schema check passes. For models whose Valid% drops below 50% we do not report the remaining strategy metrics, because they would be dominated by a small and biased subset of samples.

**ViolSev** and **LowCtrl**% quantify compliance and role adherence under our offline validator. For each schema-valid policy $y$, we apply the layered validator described in Section 3.5 and obtain its discrete violation severity level $L(y) \in \{0, \dots, 4\}$ (**Table 2**), where larger values indicate violations of higher-priority constraint layers. We report ViolSev as the dataset average of $L(y)$ over schema-valid policies (lower is better). In parallel, we check whether $y$ contains any impermissible low-level vehicle control commands (e.g., explicit throttle, brake, or steering adjustments). We report the proportion of such cases among schema-valid policies as LowCtrl\%.

**HazF1** measures whether the policy addresses decision-critical hazards implied by the perceived state. For each prompt $u = (z, p^{drv}, p^{veh}, k)$, we deterministically derive a set of key hazards $\mathcal{H}(u)$ from the joint perception output $z$ and retrieved constraints $k$. We also extract a set of hazards $\widehat{\mathcal{H}}(u, y)$ that the policy explicitly addresses via its constraints , rationale , or action descriptions. We then compute hazard precision and recall:

$$P_{haz} = \frac{|\mathcal{H}(u) \cap \widehat{\mathcal{H}}(u,y)|}{|\widehat{\mathcal{H}}(u,y)| + \epsilon}, R_{haz} = \frac{|\mathcal{H}(u) \cap \widehat{\mathcal{H}}(u,y)|}{|\mathcal{H}(u)| + \epsilon}. \quad (17)$$

We report $\text{HazF1} = 100 \times \frac{2P_{haz}R_{haz}}{P_{haz}+R_{haz}+\epsilon}$, averaged over schema-valid policies (higher is better).

Finally, to avoid homologous assessment, two independent human-expert raters (blinded to model identity) subjectively audit a stratified subset of schema-valid policies with three binary items: **No − violation** (no explicit violation of constraints in $k$ or actuator availability in $p^{veh}$), **Safe** (no obvious safety risk given $z$ and $k$), and **Evidence − supported** (key actions are supported by the cited contextual evidence from $z$ and/or $k$). Disagreements are conservatively counted as negative. We aggregate them as:

$$\text{HAS}(u,y) = 0.5\,\text{NoViol}(u,y) + 0.3\,\text{Safe}(u,y) + 0.2\,\text{EvdSup}(u,y). \quad (18)$$

where each term is binary (0 or 1) and the weights sum to 1, so $\text{HAS}(u,y) \in [0,1]$. We report $100 \times \text{HAS}$ averages over schema-valid policies. For each dataset, we evaluate HAS on a stratified subset of test samples with fixed prompt IDs across all methods and random seeds. We stratify samples into four mutually exclusive scenario groups (driver-critical, env-critical, interaction-critical, nominal) based on driver-state criticality and scene criticality; formal definitions are given in Section 4.5.3. We sample uniformly per group and report mean ± std over five seeds by auditing outputs from each independently trained run. On the stratified samples, ECPO($u$,$y$) is positively correlated with HAS, with a significantly high Spearman correlation (0.87, $p = 0.002$), suggesting the validator score is reasonably calibrated to human judgments.

### *4.4 Performance of the independent agent*

In this section, we evaluate the independent perception-strategy pipeline. Notes that, for zero-shot evaluation, we use the same task-specific instruction as in SFT.

#### 4.4.1 Perception agent results

Across all three datasets, SFT yields substantially larger gains than swapping MLLM backbones. In zero-shot mode, open-source models often underperform on schema-constrained multi-label heads and three-stage summaries. After SFT, all 7--8B backbones become competitive with (and often exceed) the GPT-5 baseline while remaining deployable for in-cabin settings.

*DriveLM*. **Table 3** summarizes DriveLM results. For multi-object recognition, the best zero-shot backbone (InternVL) reaches 34.79 IoU, 13.42% EMR, and 46.03% F1, while MILD(InternVL) improves them to 63.12, 60.27%, and 73.84%, respectively. This large jump (especially on EMR) indicates that schema-constrained SFT substantially improves consistency in set prediction. For the Initial/Transition /Final scene summaries, BertScore is already high in zero-shot mode, but SFT consistently lifts it to the low-90s across stages. N-gram metrics show the clearest change: BLEU moves from low single digits to above 20, and ROUGE rises from roughly 20 to the low/mid 40s. Overall, the fine-tuned perception agent produces more faithful entity-relation descriptions that better support downstream strategy reasoning.

*DMD*.

**Table 4** reports in-cabin results on DMD. Distraction is the hardest head in zero-shot mode: GPT-5 achieves 21.66 IoU / 31.21% F1, and LLaVA is near-zero, whereas SFT lifts InternVL/Qwen to ~61 IoU and ~63 F1. For Wheel, zero-shot GPT-5/InternVL are already strong (mid-80s in F1), and SFT further pushes performance into the low-90s. For Gaze, Qwen is strong even zero-shot, and SFT mainly narrows the gap for weaker backbones, with all fine-tuned variants converging to high and stable performance. For the three-stage summaries, BertScore is high even without SFT, but SFT consistently improves faithfulness under the schema: BLEU increases substantially (from a few points to roughly 9--14), and ROUGE also improves across stages/backbones. These gains suggest better alignment to the structured template and easier downstream consumption by the strategy agent.

*AIDE*. AIDE requires joint in-/out-of-cabin inference (**Table 5**). Zero-shot GPT-5 is moderate on the four discrete heads (roughly 30-40% Acc/F1), while zero-shot open-source backbones are near-chance (mostly $< 5\%$). After SFT, all backbones generalize well: accuracies rise to the 70s--80s across tasks, with the best head varying by backbone (e.g., LLaVA is strongest on Traffic Scene accuracy, while Qwen attains the best F1 on scene/motion). For structured summaries, SFT again yields consistent improvements across stages: BertScore moves into the low-90s, and BLEU/ROUGE increase markedly. Together, these results show that a unified structure can support reliable joint driver-environment perception once the model is tuned on schema-aligned supervision.

#### 4.4.2 Strategy agent results

We evaluate the strategy head on Mixed and AIDE using offline auditable metrics (Valid%, ViolSev, LowCtrl%, HazF1) together with a human-audited HAS (**Table 6**). Training pairs $(u, y)$ are built from ground-truth perception $z^{\star}$. At test time, we construct $u$ from the predictions $z$ of the best perception agent (fine-tuned InternVL) to reflect realistic upstream noise. For zero-shot Qwen, Valid% is $< 50\%$ on both datasets, so other metrics are reported as $N/A$. For HAS, we audit outputs from each of the five independently trained runs and report mean ± std across seeds.

*Mixed*. On Mixed, zero-shot GPT-5 and LLaMA achieve 61.35% and 96.31% Valid%, while Qwen is degenerate (**Table 6**). After SFT, both LLaMA and Qwen reach ~ 96% validity, and ECPO further improves validity to $\geq 97\%$. ECPO yields the clearest offline compliance and role-adherence gains: Low-level control suggestions drop from double digits to below 1% for both backbones, and violation severity is substantially reduced. Hazard coverage also improves under ECPO, and HAS increases accordingly, indicating better human-audited compliance, safety plausibility, and evidence support under the same schema interface.

*AIDE*. AIDE shows a similar but sharper trend (**Table 6**). Zero-shot GPT-5 has moderate validity, and Qwen again fails, while std, Bert denotes BertScore, and the best results are **bold**.

**Table 3.** Perception agent performance on the DriveLM dataset. In this and following tables, values are in the form of mean ±

| Task | Metric | Zero-shot | | | | SFT | | |
|---|---|---|---|---|---|---|---|---|
| | | GPT-5 | LLaVA | InternVL | Qwen | MILD-LLaVA | MILD-InternVL | MILD-Qwen |
| Object Recognition | IoU ↑ | 26.06 ± 1.16 | 14.25 ± 2.15 | 34.79 ± 1.04 | 23.66 ± 1.18 | 43.28 ± 0.95 | 63.12 ± 1.26 | 61.34 ± 1.22 |
| | EMR ↑ | 0.85 ± 0.98 | 0.12 ± 0.19 | 13.42 ± 1.10 | 6.01 ± 0.77 | 39.24 ± 1.73 | 60.27 ± 1.38 | 59.05 ± 1.54 |
| | F1 ↑ | 38.72 ± 1.39 | 22.40 ± 2.30 | 46.03 ± 1.09 | 31.96 ± 1.46 | 51.17 ± 1.31 | 73.84 ± 1.38 | 72.35 ± 1.20 |
| Initial | Bert ↑ | 86.34 ± 0.08 | 82.42 ± 0.24 | 87.90 ± 0.09 | 86.80 ± 0.54 | 91.70 ± 0.23 | 92.14 ± 0.32 | 92.07 ± 0.12 |
| | BLEU ↑ | 4.29 ± 0.17 | 2.15 ± 0.17 | 4.05 ± 0.22 | 3.20 ± 0.18 | 24.83 ± 0.62 | 26.98 ± 1.07 | 26.45 ± 0.65 |
| | Rouge ↑ | 23.14 ± 0.23 | 21.45 ± 0.57 | 25.72 ± 0.34 | 24.30 ± 0.34 | 40.19 ± 0.54 | 42.33 ± 1.20 | 41.48 ± 0.58 |
| Transition | Bert ↑ | 85.93 ± 0.08 | 84.88 ± 0.23 | 87.90 ± 0.10 | 87.01 ± 0.53 | 90.49 ± 0.33 | 91.22 ± 0.35 | 90.96 ± 0.13 |
| | BLEU ↑ | 3.83 ± 0.15 | 1.76 ± 0.14 | 3.41 ± 0.18 | 4.23 ± 0.20 | 20.19 ± 0.65 | 22.04 ± 1.07 | 21.51 ± 0.75 |
| | Rouge ↑ | 20.46 ± 0.24 | 18.34 ± 0.49 | 23.65 ± 0.36 | 23.26 ± 0.37 | 38.28 ± 0.66 | 39.86 ± 0.93 | 39.17 ± 0.67 |
| Final | Bert ↑ | 85.13 ± 0.07 | 80.12 ± 0.73 | 87.36 ± 0.10 | 86.79 ± 0.51 | 91.15 ± 0.27 | 91.73 ± 0.40 | 91.46 ± 0.16 |
| | BLEU ↑ | 2.32 ± 0.09 | 1.15 ± 0.09 | 3.71 ± 0.15 | 5.27 ± 0.19 | 25.93 ± 0.81 | 27.42 ± 1.12 | 26.98 ± 0.88 |
| | Rouge ↑ | 16.55 ± 0.19 | 15.22 ± 0.34 | 22.73 ± 0.30 | 23.82 ± 0.32 | 45.45 ± 0.83 | 47.18 ± 1.14 | 46.50 ± 0.85 |

**Table 4.** Perception agent performance on the DMD dataset.

| Task | Metric | Zero-shot | | | | SFT | | |
|---|---|---|---|---|---|---|---|---|
| | | GPT-5 | LLaVA | InternVL | Qwen | MILD-LLaVA | MILD-InternVL | MILD-Qwen |
| Distraction | IoU ↑ | 21.66 ± 4.50 | 1.58 ± 1.54 | 20.20 ± 3.85 | 19.94 ± 4.22 | 48.21 ± 4.19 | 61.08 ± 1.66 | 60.95 ± 1.09 |
| | EMR ↑ | 14.41 ± 4.28 | 0.90 ± 1.13 | 14.41 ± 4.73 | 12.16 ± 4.28 | 33.96 ± 4.51 | 47.88 ± 1.95 | 47.12 ± 4.73 |
| | F1 ↑ | 31.21 ± 4.40 | 3.03 ± 3.16 | 30.25 ± 4.96 | 31.81 ± 4.66 | 58.72 ± 5.73 | 63.15 ± 1.83 | 62.70 ± 0.40 |
| Wheel | IoU ↑ | 70.99 ± 4.04 | 1.47 ± 1.10 | 75.17 ± 4.53 | 68.45 ± 4.97 | 68.81 ± 4.86 | 83.17 ± 1.62 | 80.24 ± 3.76 |
| | EMR ↑ | 67.12 ± 5.43 | 0.68 ± 0.85 | 84.07 ± 4.07 | 71.53 ± 5.09 | 72.88 ± 5.09 | 90.42 ± 1.32 | 89.83 ± 3.56 |
| | F1 ↑ | 83.81 ± 2.49 | 3.95 ± 2.95 | 86.32 ± 2.70 | 81.62 ± 2.75 | 81.27 ± 3.67 | 93.05 ± 1.18 | 92.31 ± 2.23 |
| Gaze | IoU ↑ | 58.38 ± 3.70 | 10.15 ± 1.65 | 30.36 ± 4.55 | 52.84 ± 3.45 | 71.40 ± 4.58 | 79.44 ± 2.32 | 82.65 ± 3.72 |
| | EMR ↑ | 35.09 ± 5.28 | 4.20 ± 1.10 | 27.64 ± 4.66 | 51.68 ± 4.35 | 68.63 ± 4.81 | 78.23 ± 2.32 | 81.06 ± 4.19 |
| | F1 ↑ | 68.36 ± 2.80 | 15.30 ± 2.10 | 32.88 ± 3.92 | 57.02 ± 2.61 | 79.71 ± 3.62 | 84.56 ± 2.22 | 87.01 ± 3.16 |
| Initial | Bert ↑ | 87.85 ± 0.15 | 79.34 ± 0.22 | 88.78 ± 0.17 | 88.62 ± 0.19 | 90.52 ± 0.27 | 91.03 ± 0.32 | 90.95 ± 0.24 |
| | BLEU ↑ | 5.22 ± 0.46 | 3.45 ± 0.23 | 4.87 ± 0.39 | 5.90 ± 0.47 | 12.92 ± 1.38 | 14.22 ± 1.18 | 13.94 ± 1.43 |
| | Rouge ↑ | 36.15 ± 1.02 | 33.78 ± 0.63 | 35.66 ± 1.11 | 27.11 ± 0.76 | 35.97 ± 1.46 | 38.15 ± 1.27 | 37.72 ± 1.43 |
| Transition | Bert ↑ | 86.75 ± 0.13 | 80.12 ± 0.74 | 88.43 ± 0.16 | 88.46 ± 0.19 | 88.98 ± 0.30 | 89.58 ± 0.44 | 89.43 ± 0.34 |
| | BLEU ↑ | 3.05 ± 0.20 | 2.05 ± 0.13 | 3.91 ± 0.35 | 4.16 ± 0.35 | 9.18 ± 1.61 | 11.47 ± 1.27 | 11.19 ± 1.89 |
| | Rouge ↑ | 25.31 ± 0.97 | 23.88 ± 0.63 | 30.50 ± 1.11 | 24.26 ± 0.91 | 27.26 ± 1.71 | 30.12 ± 1.54 | 29.43 ± 2.11 |
| Final | Bert ↑ | 88.17 ± 0.15 | 82.95 ± 0.23 | 89.49 ± 0.15 | 89.16 ± 0.14 | 90.33 ± 0.20 | 90.79 ± 0.38 | 90.66 ± 0.22 |
| | BLEU ↑ | 4.74 ± 0.31 | 3.18 ± 0.20 | 4.30 ± 0.37 | 5.75 ± 0.49 | 9.93 ± 0.95 | 10.88 ± 1.16 | 10.65 ± 1.09 |
| | Rouge ↑ | 34.44 ± 0.78 | 31.65 ± 0.53 | 34.33 ± 0.86 | 24.83 ± 0.65 | 30.97 ± 1.09 | 32.84 ± 1.28 | 32.25 ± 1.11 |

**Table 5.** Perception agent performance on the AIDE dataset.

| Task | Metric | Zero-shot | | | | SFT | | |
|---|---|---|---|---|---|---|---|---|
| | | GPT-5 | LLaVA | InternVL | Qwen | MILD-LLaVA | MILD-InternVL | MILD-Qwen |
| Emotion | Acc ↑ | 35.42 ± 3.27 | 2.07 ± 1.13 | 4.12 ± 1.07 | 3.55 ± 1.10 | 52.41 ± 3.80 | 75.43 ± 3.24 | 78.10 ± 3.80 |
| | F1 ↑ | 29.85 ± 3.41 | 1.88 ± 0.43 | 3.85 ± 1.07 | 3.22 ± 1.07 | 50.54 ± 3.88 | 69.22 ± 2.70 | 71.84 ± 3.89 |
| Behavior | Acc ↑ | 38.12 ± 2.24 | 0.25 ± 0.31 | 0.58 ± 0.43 | 0.45 ± 0.40 | 56.38 ± 3.88 | 85.67 ± 3.29 | 88.10 ± 4.14 |
| | F1 ↑ | 31.45 ± 3.33 | 0.21 ± 0.28 | 0.62 ± 0.44 | 0.48 ± 0.40 | 53.75 ± 3.71 | 87.14 ± 2.92 | 89.58 ± 4.12 |
| Traffic Scene | Acc ↑ | 42.66 ± 3.51 | 0.18 ± 0.23 | 0.35 ± 0.39 | 0.25 ± 0.31 | 79.83 ± 3.28 | 78.56 ± 3.44 | 74.11 ± 4.75 |
| | F1 ↑ | 40.25 ± 2.70 | 0.16 ± 0.21 | 0.32 ± 0.38 | 0.22 ± 0.29 | 77.22 ± 3.28 | 83.42 ± 2.66 | 85.69 ± 3.02 |
| Vehicle Motion | Acc ↑ | 30.88 ± 3.43 | 0.35 ± 0.41 | 0.68 ± 0.56 | 0.52 ± 0.47 | 63.45 ± 4.40 | 76.28 ± 2.71 | 79.39 ± 5.18 |
| | F1 ↑ | 28.15 ± 2.27 | 0.31 ± 0.39 | 0.65 ± 0.55 | 0.48 ± 0.46 | 60.68 ± 3.97 | 74.88 ± 3.24 | 77.59 ± 3.71 |
| Initial | Bert ↑ | 88.97 ± 0.14 | 85.64 ± 0.38 | 90.31 ± 0.14 | 89.62 ± 0.16 | 92.37 ± 0.23 | 92.74 ± 0.41 | 92.60 ± 0.24 |
| | BLEU ↑ | 4.59 ± 0.27 | 1.94 ± 0.24 | 8.14 ± 0.52 | 7.67 ± 0.47 | 23.17 ± 1.51 | 25.13 ± 1.25 | 24.98 ± 1.59 |
| | Rouge ↑ | 32.05 ± 0.89 | 16.90 ± 1.06 | 37.39 ± 0.88 | 34.48 ± 0.88 | 52.90 ± 1.48 | 50.18 ± 1.47 | 49.22 ± 1.53 |
| Transition | Bert ↑ | 87.40 ± 0.12 | 84.79 ± 0.09 | 88.85 ± 0.15 | 88.69 ± 0.14 | 90.20 ± 0.20 | 90.35 ± 0.44 | 90.19 ± 0.20 |
| | BLEU ↑ | 3.37 ± 0.20 | 1.70 ± 0.35 | 4.56 ± 0.36 | 4.50 ± 0.33 | 10.98 ± 1.02 | 11.22 ± 1.30 | 10.48 ± 1.06 |
| | Rouge ↑ | 19.28 ± 0.72 | 13.74 ± 0.82 | 23.63 ± 1.03 | 24.02 ± 0.88 | 35.17 ± 1.40 | 32.14 ± 1.48 | 30.98 ± 1.40 |
| Final | Bert ↑ | 88.62 ± 0.12 | 85.72 ± 0.35 | 89.75 ± 0.13 | 89.48 ± 0.13 | 90.27 ± 0.15 | 90.52 ± 0.41 | 90.48 ± 0.14 |
| | BLEU ↑ | 3.74 ± 0.19 | 1.55 ± 0.12 | 4.76 ± 0.31 | 4.92 ± 0.35 | 8.16 ± 0.57 | 9.03 ± 1.16 | 8.55 ± 0.58 |
| | Rouge ↑ | 26.36 ± 0.78 | 16.59 ± 0.90 | 31.21 ± 0.77 | 29.11 ± 0.86 | 37.93 ± 0.98 | 31.47 ± 1.42 | 30.23 ± 0.81 |

LLaMA is valid but only moderately aligned. After SFT, validity becomes near-perfect for both backbones, and ECPO reaches 100% validity. Safety differences are driven mainly by low-level control. Zero-shot LLaMA frequently issues impermissible control commands, whereas ECPO suppresses this to around 1% (LLaMA) and below 0.5% (Qwen). Hazard coverage and HAS also improved substantially under ECPO, indicating hazard-aware and constraint-consistent strategies as confirmed by human audit.

In summary, zero-shot LLMs are unreliable for safety-sensitive in-cabin advisory policies under a strict schema. SFT is necessary for stable schema validity, while ECPO further reduces low-level control overrides and improves hazard-grounded, constraint-consistent strategies under the same interface.

### *4.5 Ablation Study*

#### 4.5.1 Ablation on ECPO design

**Table 7** ablates preference objectives for the Qwen-based strategy agent. We compare standard SFT, DPO, and several ECPO variants that drop one of the three reward terms ($s_{core}$, $s_{evd}$, $s_{str}$). On Mixed, DPO yields only marginal changes over SFT. Removing the structural term $s_{str}$ causes a clear drop in schema validity (Valid%), confirming that structure-aware rewards are necessary for schema-conforming actions. Removing the safety core term $s_{core}$ increases HazF1 but leaves noticeably higher ViolSev/LowCtrl, indicating that coverage alone does not prevent unsafe or role-violating outputs. Removing the evidence term $s_{evd}$ strongly improves safety indicators but reduces hazard coverage on Mixed. Full ECPO achieves the best overall trade-off across validity, safety, and hazard coverage. On AIDE, validity is saturated for most objectives except when dropping $s_{str}$, which again hurts schema compliance. DPO improves safety/grounding-related metrics but slightly reduces Valid% compared with SFT. Notably, removing $s_{evd}$ does not collapse HazF1 on AIDE, but its HAS remains clearly lower than full ECPO, suggesting that the three reward components are complementary and that optimizing them jointly yields the most consistently aligned behavior across datasets under human audit.

Besides, we also examine the optimization stability of the strategy agent with or without the ECPO optimization term. As shown in **Fig. 4.** Training loss curves of the strategy agent with and without ECPO on AIDE., although the ECPO term brings slightly higher loss value, the training converges smoothly in practice. This suggests that the supervised objective already provides a sufficient optimization anchor, while the ECPO term acts as an auxiliary preference regularizer rather than destabilizing policy updates.

### 4.5.2 Ablation on training data size

**Fig. 5.** Performance of Qwen-based strategy agent with different sizes of ECPO training data. The x-axis shows the percentage of the training set used for ECPO training. The y-axis of each subplot corresponds to the metric indicated in its title: Valid%, ViolSev, LowCtrl%, HazF1, and HAS studies data scaling for ECPO training with the Qwen backbone. Increasing the amount of strategy training data yields smooth and monotonic improvements on both Mixed and AIDE. Valid% rises rapidly with modest data, while ViolSev and LowCtrl% decrease steadily as training grows. AIDE is more sample-efficient and saturates earlier, consistent with its more structured supervision and simpler action space. Hazard coverage (HazF1) and the human-audited HAS improve more gradually, but follow the same monotonic trend. Finally, ECPO fine-tuning consistently provides a small additional gain over the strongest SFT checkpoint at the same data scale, acting as a targeted ``polisher'' that improves safety and grounding without requiring extra supervised labels.

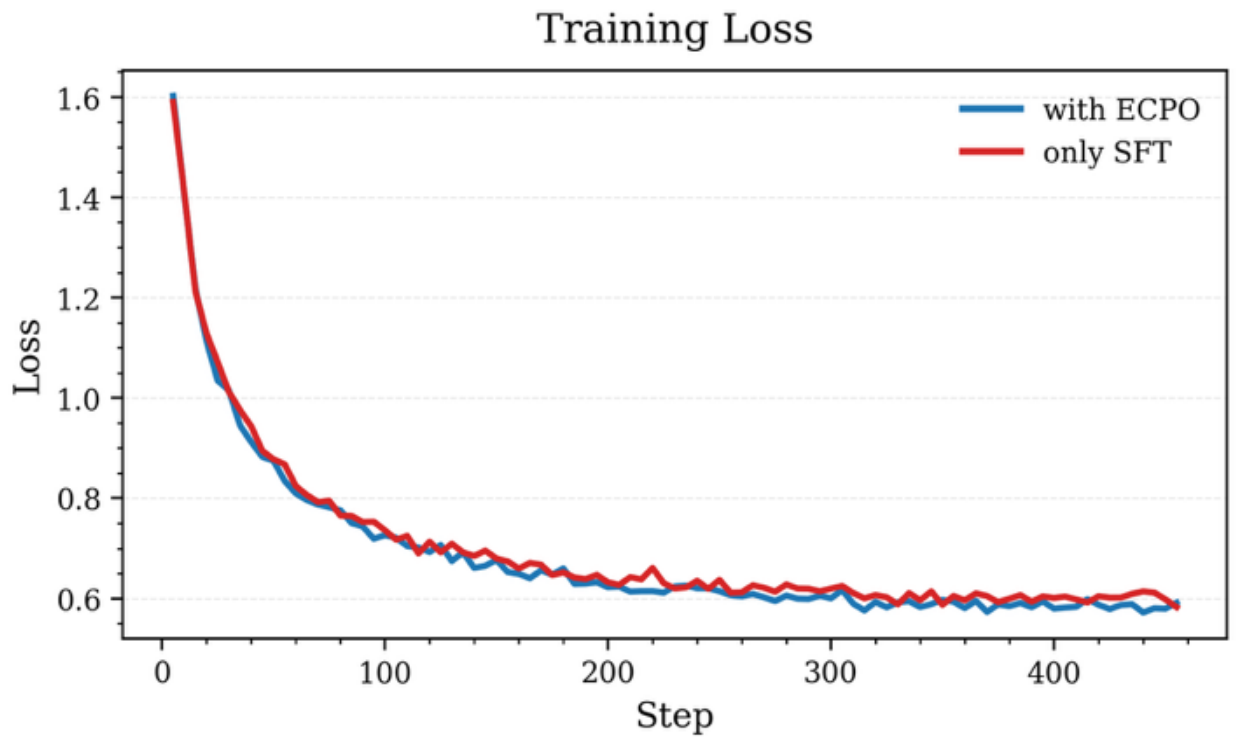

**Fig. 4.** Training loss curves of the strategy agent with and without ECPO on AIDE.

### 4.5.3 Ablation of joint in-/out-of-cabin perception

One of the key claims of MILD is that the strategy agent should reason over a joint view of the driver and the surrounding environment. To test this, we take the AIDE test set and group samples into four scenario categories using the ground-truth discrete labels for Emotion, Behavior, Traffic Scene, and Vehicle Motion: driver-critical (any non-nominal driver emotion or behavior while the traffic scene and vehicle motion are nominal), env-critical (nominal driver but critical traffic scene or vehicle motion), interaction-critical (both driver and environment critical), and nominal (all four heads nominal). For each scenario, we evaluate three variants of the Qwen-based ECPO strategy agent: (i) **Full**, which uses the full perception summary $z$ (driver $x^{in}$ and environment $x^{out}$); (ii) **w/o $x^{in}$**, where all in-cabin descriptors are masked; and (iii) **w/o $x^{out}$**, where out-of-cabin descriptors are masked. The strategy head parameters are fixed across all variants.

In addition to the ECPO-based metrics, for each variant, we collect human ratings from twenty drivers (5 males, age ranged from 23 to 42), recruited in China, with at least one year of ADS driving experience. All participants provided informed consent before evaluation. The Hong Kong University of Science and Technology (Guangzhou)'s Human and Artefacts Research Ethics Committee approved this study (protocol number: HKUST(GZ)-HSP-2025-0120). We use a within-subject design: each participant rates randomized policies from all variants, blinded to model identity. Each policy is evaluated on three 5-point Likert scales. **Adequacy** and **Comfort** are adapted from the ``usefulness'' and ``satisfaction'' factors of (Van Der Laan et al., 1997), capturing how appropriate and how comfortable the suggested strategy feels in context. **Explanation** follows the explanation-understanding items used in (Koo et al., 2015), measuring how clearly the policy states its rationale and constraints. We quantify rating consistency using an intraclass correlation coefficient with absolute agreement, reporting the average-measures ICC over twenty raters. We obtain high inter-rater reliability: $\mathrm{ICC}_{\mathrm{Adequacy}} = 0.91$ (95% CI: [0.87, 0.95]), $\mathrm{ICC}_{\mathrm{Comfort}} = 0.92$ (95% CI: [0.90, 0.94]), and $\mathrm{ICC}_{\mathrm{Explanation}} = 0.90$ (95% CI: [0.86, 0.94]).

As shown in

**Table 8**, across the three non-nominal scenarios, the **Full** variant consistently achieves the highest ECPO score and HazF1 while keeping Valid% near 100% and ViolSev low, and it also dominates the human ratings. In driver-critical scenes, ablating in-cabin perception (**w/o $x^{in}$**) leads to the largest drop in both ECPO score (from 90.15 to 76.40) and human judgments: Adequacy falls from 4.52 to 2.85, Comfort from 4.45 to 2.92, and Explanation from 4.75 to 2.42. In env-critical scenes, removing $x^{out}$ causes the most severe degradation, indicating that the agent can no longer account for external hazards and traffic dynamics. Interaction-critical scenes show the strongest synergy: either single-modality variant yields ECPO scores around 78 and Adequacy close to 3.0, whereas the Full model reaches 91.8 ECPO with Adequacy, Comfort, and Explanation all around 4.8, suggesting that many high-risk situations arise from the interaction between driver state and environment and cannot be resolved by attending to only one side.

For nominal scenes, all three variants perform similarly well on both ECPO and human scores (Adequacy and Comfort above 4.1 for all variants), with the Full model providing only a modest advantage in HazF1 and Explanation. This pattern is desirable: MILD does not introduce unnecessary intervention in easy cases, but when either the driver or the environment becomes critical, joint in-/out-of-cabin perception improves both hazard-awareness and constraint-compliance metrics (e.g., HazF1 and ECPO- based validator scores) and human-judged strategy quality.

**Table 6.** Strategy agent performance on the Mixed and AIDE datasets. "N/A" indicates metrics that are not reported because the corresponding model produces less than 50% valid policies.

| Dataset | Metric | Zero-shot | | | SFT | | ECPO | |
|---|---|---|---|---|---|---|---|---|
| | | GPT-5 | LLaMA | Qwen | MILD-LLaMA | MILD-Qwen | MILD-LLaMA | MILD-Qwen |
| Mixed | Valid% ↑ | 61.35 ± 4.12 | 96.31 ± 1.37 | 0.29 ± 0.31 | 96.46 ± 1.29 | 96.46 ± 1.34 | 97.20 ± 0.98 | 98.24 ± 1.21 |
| | ViolSev ↓ | 1.56 ± 0.33 | 1.90 ± 0.41 | N/A | 0.50 ± 0.18 | 1.92 ± 0.39 | 0.21 ± 0.12 | 0.54 ± 0.21 |
| | LowCtrl% ↓ | 10.76 ± 2.47 | 13.15 ± 2.86 | N/A | 10.06 ± 1.74 | 11.77 ± 2.11 | 0.92 ± 0.63 | 0.98 ± 0.89 |
| | HazF1 ↑ | 55.13 ± 6.32 | 55.54 ± 5.87 | N/A | 49.96 ± 5.24 | 55.71 ± 6.05 | 61.25 ± 4.98 | 64.55 ± 6.73 |
| | HAS ↑ | 69.96 ± 4.59 | 66.64 ± 4.21 | N/A | 78.91 ± 3.88 | 66.55 ± 4.37 | 80.27 ± 3.51 | 78.68 ± 3.96 |
| AIDE | Valid% ↑ | 55.25 ± 5.28 | 94.83 ± 2.03 | 0.10 ± 0.11 | 99.83 ± 0.37 | 99.83 ± 0.39 | 100.00 ± 0.00 | 100.00 ± 0.00 |
| | ViolSev ↓ | 0.40 ± 0.19 | 0.53 ± 0.15 | N/A | 0.47 ± 0.18 | 0.43 ± 0.16 | 0.43 ± 0.17 | 0.31 ± 0.14 |
| | LowCtrl% ↓ | 2.24 ± 0.91 | 12.18 ± 0.87 | N/A | 8.21 ± 0.63 | 8.04 ± 0.57 | 1.03 ± 0.49 | 0.35 ± 0.31 |
| | HazF1 ↑ | 61.73 ± 3.12 | 69.65 ± 3.48 | N/A | 61.27 ± 2.86 | 62.95 ± 3.21 | 72.04 ± 3.07 | 73.45 ± 4.95 |
| | HAS ↑ | 60.87 ± 4.03 | 63.47 ± 3.76 | N/A | 67.72 ± 3.12 | 68.36 ± 3.28 | 88.35 ± 3.19 | 88.78 ± 3.41 |

**Table 7.** Ablation of ECPO design for the Qwen-based strategy agent.

| Dataset | Metric | SFT | DPO | ECPO w/o $s_{core}$ | ECPO w/o $s_{evd}$ | ECPO w/o $s_{str}$ | ECPO |
|---|---|---|---|---|---|---|---|
| Mixed | Valid% ↑ | 96.46 ± 1.34 | 95.80 ± 1.25 | 96.85 ± 1.10 | 97.15 ± 0.95 | 89.20 ± 1.55 | 98.24 ± 1.21 |
| | ViolSev ↓ | 1.92 ± 0.39 | 1.75 ± 0.35 | 1.45 ± 0.32 | 0.65 ± 0.18 | 0.72 ± 0.22 | 0.54 ± 0.21 |
| | LowCtrl% ↓ | 11.77 ± 2.11 | 10.50 ± 1.95 | 8.50 ± 1.80 | 1.55 ± 0.65 | 2.10 ± 0.85 | 0.98 ± 0.89 |
| | HazF1 ↑ | 55.71 ± 6.05 | 57.20 ± 5.80 | 63.80 ± 6.20 | 52.10 ± 5.50 | 62.50 ± 6.10 | 64.55 ± 6.73 |
| | HAS ↑ | 66.55 ± 4.37 | 68.12 ± 4.10 | 71.55 ± 4.25 | 74.80 ± 3.80 | 75.40 ± 4.15 | 78.68 ± 3.96 |
| AIDE | Valid% ↑ | 99.83 ± 0.39 | 99.50 ± 0.45 | 99.92 ± 0.15 | 100.00 ± 0.00 | 90.45 ± 0.65 | 100.00 ± 0.00 |
| | ViolSev ↓ | 0.43 ± 0.16 | 0.40 ± 0.15 | 0.38 ± 0.15 | 0.33 ± 0.12 | 0.34 ± 0.13 | 0.31 ± 0.14 |
| | LowCtrl% ↓ | 8.04 ± 0.57 | 7.50 ± 0.52 | 6.20 ± 0.48 | 0.55 ± 0.25 | 1.15 ± 0.40 | 0.35 ± 0.31 |
| | HazF1 ↑ | 62.95 ± 3.21 | 64.50 ± 3.10 | 72.10 ± 4.50 | 73.10 ± 2.90 | 71.80 ± 4.20 | 73.45 ± 4.95 |
| | HAS ↑ | 68.36 ± 3.28 | 70.15 ± 3.15 | 74.50 ± 3.40 | 80.20 ± 3.05 | 83.50 ± 3.35 | 88.78 ± 3.41 |

**Table 8.** Ablation of joint in-/out-of-cabin perception on strategy quality on the AIDE dataset. Objective scores are ECPO-based metrics. Human ratings are collected from twenty drivers on 5-Likert scale (higher is better).

| Scenario | Variant | Valid% ↑ | ViolSev ↓ | LowCtrl% ↓ | HazF1 ↑ | HAS ↑ | Adequacy ↑ | Comfort ↑ | Explanation ↑ |
|---|---|---|---|---|---|---|---|---|---|
| Driver-critical | w/o $x^{in}$ | 99.90 ± 0.15 | 0.42 ± 0.18 | 0.65 ± 0.35 | 62.50 ± 4.10 | 76.40 ± 3.85 | 2.85 ± 0.62 | 2.92 ± 0.58 | 2.42 ± 0.55 |
| | w/o $x^{out}$ | 100.00 ± 0.00 | 0.30 ± 0.14 | 0.38 ± 0.25 | 69.80 ± 4.50 | 84.50 ± 3.20 | 3.45 ± 0.55 | 3.25 ± 0.52 | 3.35 ± 0.60 |
| | Full | 100.00 ± 0.00 | 0.28 ± 0.12 | 0.32 ± 0.20 | 76.20 ± 5.10 | 90.15 ± 3.10 | 4.52 ± 0.45 | 4.45 ± 0.42 | 4.75 ± 0.38 |
| Env-critical | w/o $x^{in}$ | 100.00 ± 0.00 | 0.30 ± 0.15 | 0.48 ± 0.30 | 71.20 ± 4.80 | 87.60 ± 3.40 | 3.95 ± 0.52 | 3.65 ± 0.55 | 3.60 ± 0.58 |
| | w/o $x^{out}$ | 99.85 ± 0.25 | 0.48 ± 0.20 | 0.75 ± 0.45 | 58.40 ± 3.50 | 74.20 ± 4.10 | 2.20 ± 0.65 | 2.50 ± 0.60 | 2.00 ± 0.55 |
| | Full | 100.00 ± 0.00 | 0.28 ± 0.13 | 0.45 ± 0.25 | 75.10 ± 5.20 | 89.50 ± 3.30 | 4.60 ± 0.40 | 4.50 ± 0.45 | 4.80 ± 0.35 |
| Interaction-critical | w/o $x^{in}$ | 99.95 ± 0.10 | 0.44 ± 0.19 | 0.68 ± 0.40 | 64.20 ± 4.20 | 78.50 ± 3.90 | 3.20 ± 0.60 | 2.80 ± 0.62 | 2.80 ± 0.58 |
| | w/o $x^{out}$ | 99.90 ± 0.15 | 0.46 ± 0.21 | 0.72 ± 0.45 | 63.80 ± 4.10 | 77.90 ± 4.05 | 3.00 ± 0.65 | 2.80 ± 0.58 | 2.60 ± 0.60 |
| | Full | 100.00 ± 0.00 | 0.22 ± 0.11 | 0.35 ± 0.22 | 78.50 ± 5.50 | 91.80 ± 3.05 | 4.80 ± 0.35 | 4.60 ± 0.38 | 4.80 ± 0.30 |
| Nominal | w/o $x^{in}$ | 100.00 ± 0.00 | 0.23 ± 0.10 | 0.15 ± 0.10 | 36.50 ± 3.80 | 88.20 ± 3.10 | 4.40 ± 0.45 | 4.20 ± 0.48 | 4.00 ± 0.50 |
| | w/o $x^{out}$ | 100.00 ± 0.00 | 0.24 ± 0.11 | 0.18 ± 0.12 | 35.80 ± 3.90 | 87.90 ± 3.15 | 4.20 ± 0.42 | 4.20 ± 0.45 | 4.00 ± 0.52 |
| | Full | 100.00 ± 0.00 | 0.20 ± 0.09 | 0.12 ± 0.08 | 40.50 ± 4.20 | 90.50 ± 2.95 | 4.50 ± 0.40 | 4.40 ± 0.38 | 4.20 ± 0.42 |

4.5.4 Sensitivity to hyperparameters

We further study the sensitivity of ECPO to two key hyperparameters: the aggregation weights ($s_{core}, s_{evd}, s_{str}$) and the candidate policy number $K$. For the aggregation weights, we vary the relative emphasis on core safety, evidence grounding, and structural completeness. As shown in **Table 9**, larger $s_{core}$yields lower violation severity (e.g., 0.26 under 0.7/0.15/0.15 ), whereas larger evidence/structure weights improve hazard coverage and human audit results (e.g., HazF1 = 74.27 and HAS = 90.13 under 0.4/0.3/0.3), at the cost of slightly higher ViolSev. Our default setting (0.5, 0.3, 0.2) achieves a balanced trade-off (ViolSev = 0.31, HazF1 = 73.45, HAS = 88.78)

Besides, for the generated candidate set size $K$, we vary $K \in \{2,4,8\}$ to evaluate the proposal's performance and latency (

**Table 10**). The policy quality changes only mildly across this range, indicating that ECPO is not overly sensitive to $K$. However, latency increases substantially as $K$ grows (p95 latency: 1.54s for $K = 2$, 2.18s for $K = 4$, and 3.32s for $K = 8$). Since $K = 2$achieves competitive ViolSev/HazF1/HAS while offering the best efficiency-quality trade-off, we take it as the default setting.

**Table 9**. Performance sensitivity experiment on aggregation weights.

| $s_{core}$ | $s_{evd}$ | $s_{str}$ | ViolSev↓ | HazF1↑ | HAS↑ |
|---|---|---|---|---|---|
| 0.7 | 0.15 | 0.15 | 0.26 | 71.82 | 87.15 |
| 0.6 | 0.2 | 0.2 | 0.29 | 72.94 | 88.31 |
| 0.5 | 0.3 | 0.2 | 0.31 | 73.45 | 88.78 |
| 0.5 | 0.2 | 0.3 | 0.3 | 73.12 | 89.56 |
| 0.4 | 0.3 | 0.3 | 0.33 | 74.27 | 90.13 |

**Table 10.** Performance sensitivity experiment on generated candidate policy number *K*.

| *K* | ViolSev↓ | HazF1↑ | HAS↑ | p95 latency |
|---|---|---|---|---|
| 2 | 0.31 | 73.45 | 88.78 | 1.54 |
| 4 | 0.32 | 73.61 | 88.24 | 2.18 |
| 8 | 0.31 | 73.13 | 87.67 | 3.32 |

### *4.6 Case Study*

#### 4.6.1 Effect of perception quality on strategy agent performance

Since the ECPO validator is defined relative to the perceived world state, we compare test-time prompts built from learned perception outputs $z$ versus oracle annotations $z^{\star}$ (**Table 11**). Specifically, in the standard setting, prompts $u$ are formed from the predicted outputs $z$ of the best perception agent (fine-tuned InternVL). In the oracle setting, we build $u$ from ground-truth annotations $z^{\star}$ and reevaluate the same ECPO-optimized strategy heads without retraining.

Basic safety is largely stable. Valid% remains high on Mixed and is saturated on AIDE, and both ViolSev and LowCtrl% change only slightly when switching from $z$ to $z^{\star}$. In contrast, grounding-related metrics improve under oracle perception. HazF1 and HAS increase consistently, supporting the intended coupling: better upstream perception mainly benefits hazard grounding and human-audited evidence supports, while ECPO-trained safety behavior is robust to small perception errors.

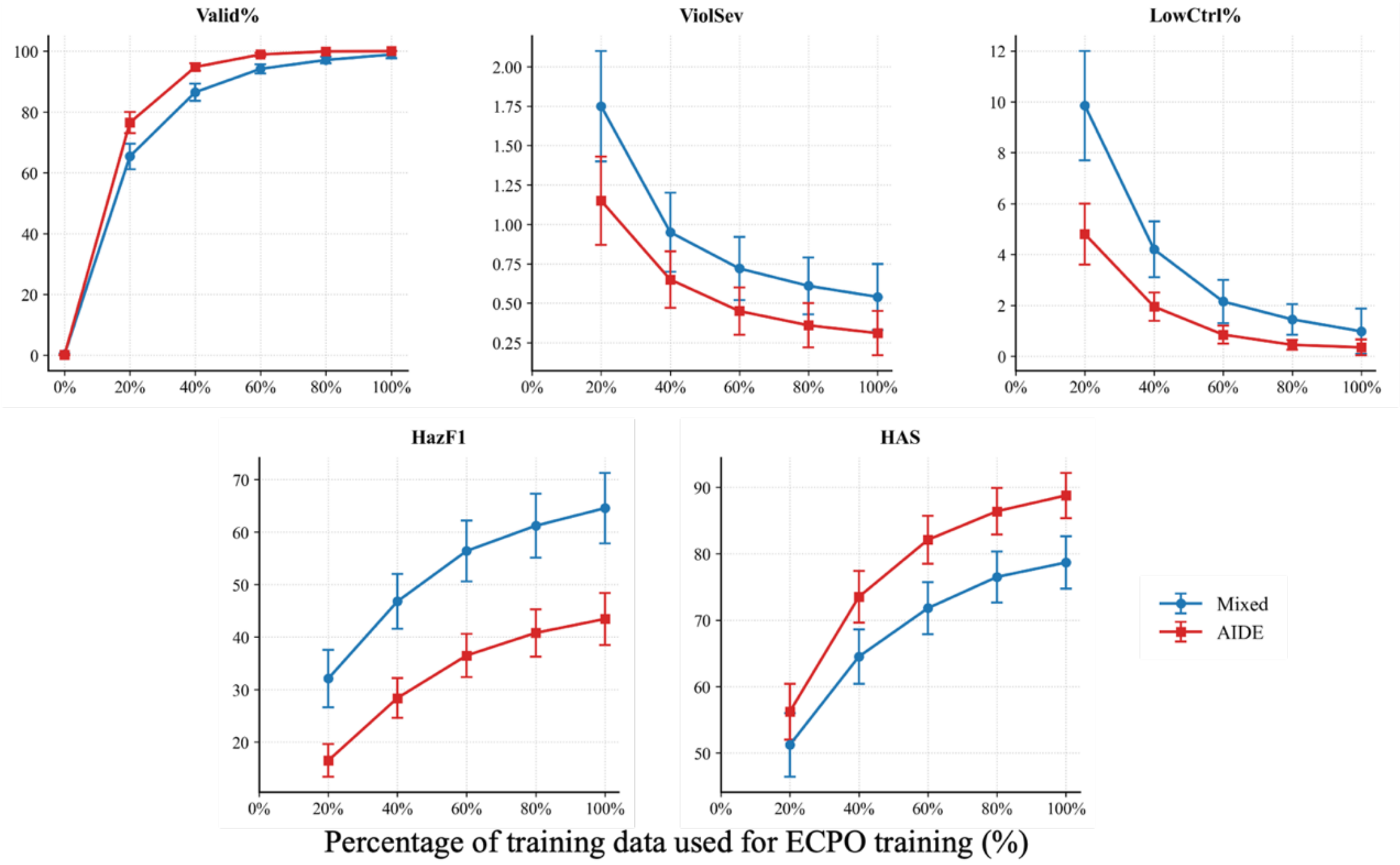

**Fig. 5.** Performance of Qwen-based strategy agent with different sizes of ECPO training data. The x-axis shows the percentage of the training set used for ECPO training. The y-axis of each subplot corresponds to the metric indicated in its title: Valid%, ViolSev, LowCtrl%, HazF1, and HAS.

#### 4.6.2 Latency-performance trade-off under RAG context and quantization

To evaluate whether MILD is compatible with practical in-cockpit deployment, we study the trade-off between performance and efficiency under two deployment-facing factors that do not modify learned parameters: (i) the amount of RAG context appended to the strategy agent (RAG context tokens), and (ii) the inference precision (Int4, Bfloat16, and Float32). For each configuration, we report p95 end-to-end inference latency and the resulting HAS, and visualize the operating points in a Pareto plot (**Fig. 6**), where lower latency and higher HAS are both preferred.

**Fig. 6** reveals two consistent trends. First, increasing the injected RAG context improves HAS up to a moderate budget, beyond which gains saturate. Adding a modest amount of context yields a clear alignment improvement across precisions, whereas further extending the prompt produces diminishing returns in HAS but continues to increase latency due to higher attention costs. Second, precision primarily determines the runtime envelope, inducing a smooth accuracy–speed trade-off. Lower precision (Int4) substantially reduces tail latency with only a limited HAS decrease relative to Bfloat16/Float32, while higher precision preserves a small alignment advantage at the cost of slower inference. Overall, these results suggest that even when the Bfloat16 strategy agent is deployed directly, the observed tail latency remains compatible with interactive, non-control cockpit mediation workloads (i.e., high-level suggestions rather than real-time vehicle actuation). This defines the intended application range of MILD: driver-facing warnings, explanation timing, assistance-mode suggestions, following-gap recommendations, and comfort/HMI adjustments that can tolerate clip-level policy refresh. In deployment, latency could be handled in a non-blocking manner: the latest validated policy remains active until a newer policy is returned, while emergency stabilization, collision avoidance, and other millisecond-level vehicle behaviors remain the responsibility of the underlying ADS stack. When tighter compute or storage budgets are required, downshifting to Int4 offers a practical way to further reduce both inference cost and model footprint while slightly sacrificing performance.

### 4.6.3 Qualitative analysis

This section presents qualitative examples to demonstrate how the MILD agent (Qwen version) translates joint perception into tailored, driver-centric strategies compared to a strong zero-shot baseline (GPT-5). **Fig. 7** illustrates two representative scenarios from the AIDE test set, highlighting key policy divergences in red. The comparison reveals that while the baseline often generates generic or profile-agnostic suggestions, MILD produces precise, multi-actuator interventions that strictly adhere to individual sensitivities, environmental context, and safety constraints.

**Table 11**. Effect of perception quality on ECPO strategy metrics. ``With $z$" uses prompts built from the learned perception agent, while ``With $z^{\star}$" uses ground-truth perception summaries.

| Dataset | Metric | With $z$ LLaMA | With $z$ Qwen | With $z\star$ LLaMA | With $z\star$ Qwen |
|---|---|---|---|---|---|
| Mixed | Valid% ↑ | 97.2 | 98.24 | 97.05 | 99.01 |
| | ViolSev ↓ | 0.21 | 0.54 | 0.2 | 0.47 |
| | LowCtrl% ↓ | 0.92 | 0.98 | 0.93 | 0.99 |
| | HazF1 ↑ | 61.25 | 64.55 | 62.88 | 65.38 |
| | HAS ↑ | 80.27 | 78.68 | 81.92 | 79.32 |
| AIDE | Valid% ↑ | 100 | 100 | 100 | 100 |
| | ViolSev ↓ | 0.43 | 0.31 | 0.38 | 0.31 |
| | LowCtrl% ↓ | 1.03 | 0.35 | 1.11 | 0.31 |
| | HazF1 ↑ | 72.04 | 73.45 | 73.56 | 73.88 |
| | HAS ↑ | 88.35 | 88.78 | 90.04 | 89.52 |

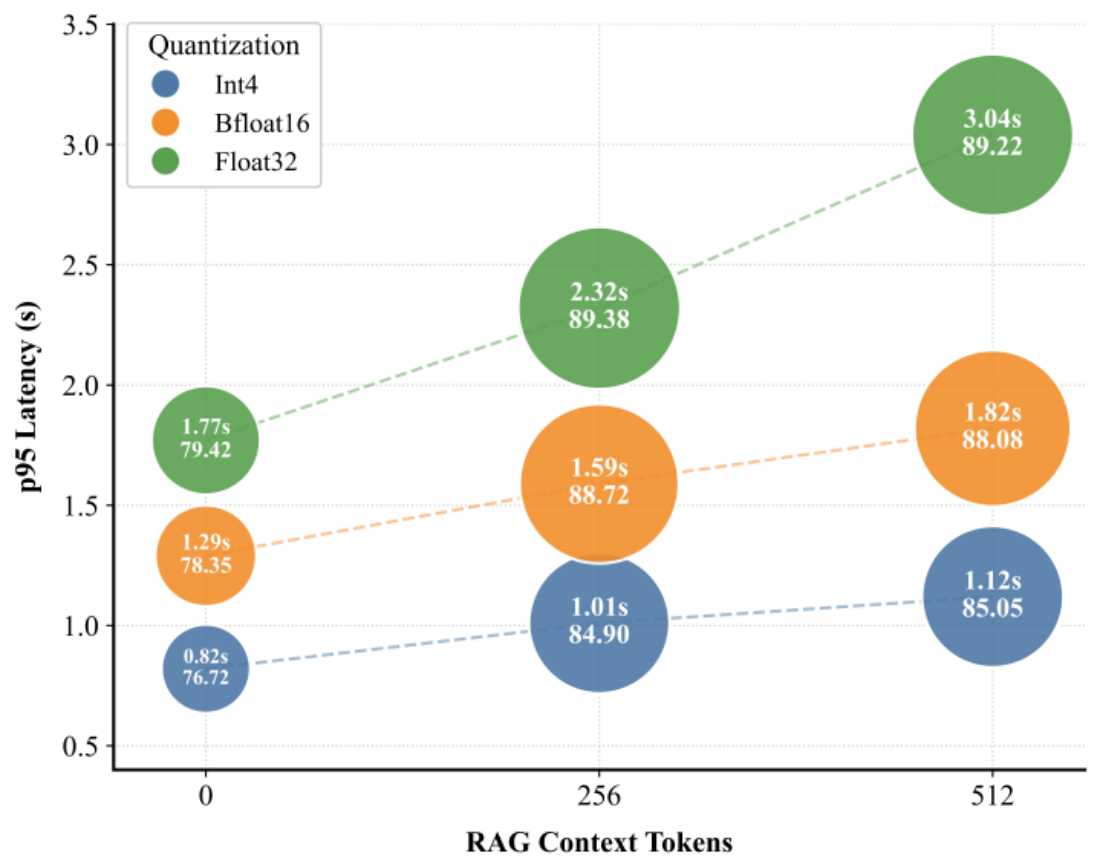

**Fig. 6.** Pareto analysis of deployment configurations for MILD. We vary the RAG context length appended to the strategy agent (x-axis) and the inference precision (color: Int4, Bfloat16, Float32), and report p95 end-to-end latency together with HAS (higher HAS means bigger marker size). Each marker is annotated with its measured latency and HAS.

In Scenario 1 (top row), the vehicle is performing a low-speed parking maneuver in dense stop-and-go traffic, and the driver shows anxiety with frequent scanning behavior. The driver profile contains several comfort-related constraints (high motion-sickness susceptibility, high noise sensitivity, and a preferred cabin temperature of $22 - 24°C$) but also indicates an active/assertive driving style preference. This creates a typical preference--context conflict: an active style is incompatible with the current tight-gap parking context and the surrounding traffic risks implied by the joint perception. The zero-shot GPT-5 policy is sparse and does not explicitly arbitrate this conflict. It proposes a generic parking-related response and sets the HVAC to $26°C$, which violates the stated comfort band, while providing little evidence-grounded justification. In contrast, MILD makes the safety envelope and conflict resolution explicit through its evidence-linked rationale: it highlights “heavy traffic” and advises to ”mind surrounding gaps” rather than encouraging aggressive progression, and simultaneously respects the hard comfort constraints by targeting $23°C$ with fresh-air intake and suppressing “audio and repeated alerts” to accommodate noise sensitivity and motion-sickness risk. This example demonstrates that MILD treats driving-style preferences as soft and condition-dependent, while enforcing evidence-consistent caution and hard comfort constraints in risk-prone contexts.

Scenario 2 (bottom row) depicts a reversing maneuver in a rainy environment (“rainy day”), where the driver appears angry and distracted. While GPT-5 characterizes the scene incorrectly as a "busy urban area" and suggests "warm amber ambient light", MILD leverages specific environmental cues to recommend a "brief safe pause" due to the rain and backing maneuver. Furthermore, recognizing the driver's anger and motion sickness risks, MILD activates fresh-air intake with face-level airflow to ease physiological stress and selects “cool-white ambient lighting.” This lighting choice is designed to support calm focus without adding glare, contrasting with GPT-5's amber setting, thus demonstrating MILD's ability to synthesize external weather conditions with internal driver states for a safer intervention.

## 5 Discussion

The results support three main claims regarding MILD. First, a unified, schema-driven approach enables compact multimodal LLMs (7–8B) to match or exceed larger GPT-series models in specialized driving tasks. Fine-tuning on DMD, DriveLM, and AIDE yields 20–30 point improvements in IoU and F1 scores over zero-shot baselines. This confirms that domain-specific schema constraints allow smaller, deployable models to bridge the gap to general-purpose frontiers, making them viable for in-cabin resource-constrained environments. This will also bring advantages for privacy protection and cybersecurity.

Second, ECPO-style preference optimization is essential for transforming LLMs into reliable, human-aligned strategy agents. While zero-shot models and standard SFT often suffer from safety violations or inappropriate low-level control, ECPO significantly suppresses these issues. By optimizing for safety, grounding, and structural coherence, ECPO produces agents that are not only compliant but also better justified. Our ablation study confirms that all three reward components are necessary: removing any one leads to a trade-off between safety and interpretability, echoing challenges in multi-objective RL.

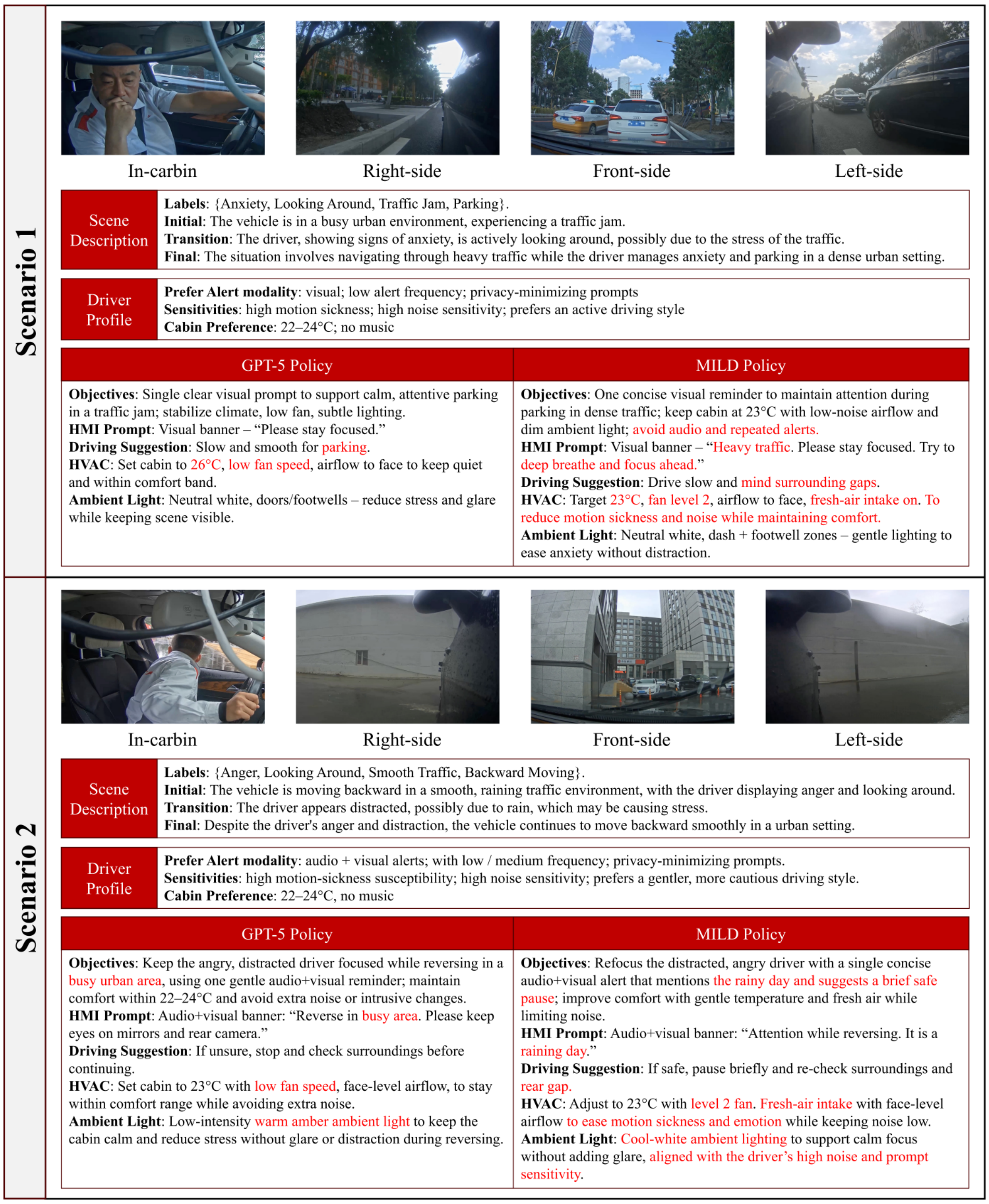

**Fig. 7.** The example output policy of zero-shot GPT-5 and MILD(Qwen) in different scenarios from AIDE. The key differences in policy are marked in red.

We also observe a notable zero-shot asymmetry among open-source backbones in AIDE. We do not interpret this as a simple capacity issue. On the perception side, all open-source MLLMs perform poorly on AIDE in zero-shot, suggesting that the main difficulty lies in the joint structured target itself, which requires simultaneous in-cabin and out-of-cabin inference under fixed discrete heads and three-stage summaries. On the strategy side, however, the failure modes differ across backbones under the same policy interface. Zero-shot LLaMA is mostly able to produce parseable PolicyAction objects, but often oversteps the intended high-level advisory role by issuing impermissible low-level control suggestions. Zero-shot Qwen, in contrast, frequently fails earlier at schema validity, i.e., it does not reliably instantiate the full policy object at all. This indicates that, under the same strict interface, different backbones may fail at different stages of structured generation. Importantly, this should not be over-read as a general weakness of Qwen: it is already strong on simpler zero-shot perception settings such as DMD Gaze, and it recovers sharply on both perception and strategy tasks after SFT/ECPO. We therefore view the main bottleneck here as backbone-specific zero-shot robustness to the bidirectional structured interface, rather than raw model capacity alone.

Furthermore, joint in-cabin and out-of-cabin perception is crucial for effective assistance in high-risk scenarios. Evidence from AIDE and human evaluations shows that ablating either perception stream significantly degrades hazard coverage (HazF1) and safety plausibility. The "Full" variant is consistently preferred by users, particularly in complex interactions where risks arise from the coupling of driver impairment and environmental hazards. This suggests that MILD's holistic context awareness improves both objective safety proxies and subjective user trust, positioning it as an effective "assistant-on-top" layer that mediates between the human, the vehicle, and the environment. More broadly, the current perception-to-strategy interface is intentionally compact rather than lossless. In MILD, the main technical focus is the strategy agent and its validator-driven alignment mechanism, while the perception agent serves as a lightweight upstream interface. We intentionally do not endow the strategy agent itself with full multimodal perception, because doing so would enlarge the policy model, increase latency, and potentially introduce inconsistencies with the perception outputs already used by the underlying driving system. Instead, the structured summary is used as a policy-facing abstraction: it may omit fine-grained visual nuances, but it enables a lighter, auditable, and more modular strategy layer. This trade-off is also reflected in Section 4.6.1, where stronger perception mainly improves grounding-related measures rather than the basic role-boundary and safety behavior of the strategy agent.

The proposed approach contributes to the emerging concept of agentic vehicles (AgVs), in which one or more AI agents endow vehicles with goal-oriented reasoning, learning, and decision-making capabilities for both driving and auxiliary functions (Yu et al., 2026). By embedding such mediator agent at the vehicle level, AgVs are expected to serve as foundational components of future agentic transportation and mobility services, with the potential to enhance individuals' healthcare (Jiyao Wang, Lu, et al., 2025; Jiyao Wang, Yang, et al., 2025; 2026), and personalized travel plan (Jiyao Wang, Zhang, et al., 2026), and adaptable iHMI (Jiyao Wang, Yan, et al., 2026).

# 6 Limitations

Despite these gains, several limitations remain. First, personalization and reward modeling remain static. MILD does not yet account for long-term driver adaptation or large-scale inter-driver variability. Additionally, while MILD acts as an advisory layer, current protocols cannot capture closed-loop dynamics or long-horizon risk accumulation. This evaluation is appropriate for the present scope of MILD, which is designed as a high-level mediation layer rather than an end-to-end controller, but it cannot capture closed-loop human adaptation, downstream vehicle dynamics, or long-horizon operational benefits. Future work should involve high-fidelity simulators and on-road studies to validate real-world safety outcomes and end-to-end deployment overheads (e.g., HMI rendering and I/O latency). Besides, we also note that the current system does not explicitly encode a human-factors intervention model that guarantees the causal efficacy of specific actuator suggestions. In this sense, such interventions are selected under preference and context constraints, while drawing on the LLMs' prior knowledge about potentially helpful regulation or comfort adjustments. Whether these specific suggestions produce measurable downstream improvements should be examined in future human-factors studies.

Second, because only AIDE provides naturally paired views, our "Mixed" dataset relies on synthetic pairing between DMD and DriveLM. Because Mixed may induce spurious driver-scene associations, it should not be interpreted as validating causal coupling between human state and external events. Although we mitigate artifacts through careful shuffling, this remains a proof-of-concept that requires further validation as more synchronized in-/out-of-cabin datasets become available.

A further limitation concerns teacher dependence and annotation subjectivity in data construction. We utilize GPT-assisted drafting followed by manual review to construct the policy references and verbalize discrete labels into our target perception agent's text format. Manual review improves schema validity, constraint consistency, and evidence grounding, but may not fully remove the teacher model's stylistic or normative preferences, such as conservative warning tendencies, verbosity patterns, or comfort-oriented interventions. Moreover, high-level HMI and comfort suggestions are not uniquely determined by a single scene; multiple advisory policies may be reasonable under the same constraints. Future work may compare alternative teacher models, increase the proportion of human-authored references, and examine inter-annotator agreement to better characterize the sensitivity of the current teacher-assisted data construction pipeline.

Finally, the current paper does not explicitly address retrieval failures or erroneous constraint retrieval in the RAG module. Handling explicit within-layer contradictions in profiles or robust retrieved knowledge is left for future work.

# 7 Conclusion

In this work, we introduced MILD, a modular driver--vehicle assistant that factorizes decision-making into a multi-modal perception agent and a text-only strategy agent coupled through a unified, schema-based representation. On top of this, we proposed ECPO, a decomposed preference-optimization framework that turns a rule- and profile-informed validator into an offline reward model over safety, evidence grounding, and structural coherence. By aligning heterogeneous datasets to a common target schema, we showed that compact open-source MLLMs can be turned into strong perception agents across in-cabin, out-of-cabin, and naturally paired settings, matching or surpassing a much larger GPT-5 baseline on discrete labels and structured summaries. Building on these structured perception outputs, ECPO-trained strategy agents achieve consistently high schema validity, markedly fewer safety and comfort violations, substantially lower rates of impermissible low-level control, and higher coverage of decision-critical hazards implied by the joint perception and constraint context, supporting stronger strategy-level policy quality under ablations, joint perception studies, and human evaluations. Our deployment-oriented analysis suggests that MILD can operate within realistic in-cockpit interaction budgets when used as a non-control mediator for high-level advisory, HMI, and comfort-related strategies.

More broadly, MILD advocates for structured, schema-based interfaces as a principled approach to integrating LLMs into intelligent vehicles. Constraining all agents to read and write in a shared structure reduces the stochasticity and parsing fragility of free-form generation, while providing a legible substrate for expressing, checking, and explaining safety rules, comfort preferences, driver profiles, and multi-actuator policies. ECPO, in turn, shows how such validators can be formulated as decomposed rewards, enabling domain-specific alignment of LLMs to safety, comfort, and role boundaries, rather than relying on end-to-end black-box controllers. Framing the language model as a mediator between the driver, automation, and environment both concretizes the vision outlined in our introduction and opens up follow-up work at the intersection of automated driving, human factors, and HMI—for example, longitudinal personalization of strategies, richer human-in-the-loop refinement of validators, and tighter coupling to low-level control stacks. We hope that the proposed architecture, schema design, and ECPO formulation will inform future research on human-centric driving agents and offer the industry a reference design pattern and empirical basis for exploring language-enabled assistants as transparent, configurable high-level overlays on existing ADS systems.

## Author contributions

**Jiyao Wang**: Conceptualization of this study, Methodology, Writing - Original draft preparation. **Yunbiao Wang**: Writing - Original draft preparation. **Yubo Jiao**: Methodology. **Xiao Yang**: Software. **Dengbo He**: Resources. **Sasan Jafarnejad**: Writing - Review Editing. **Raphaël Frank**: Supervision, Resources, Writing - Review. **Jiangbo Yu**: Supervision, Resources, Funding acquisition, Writing - Review Editing.

## Replication and data sharing

For the original data of DMD, DriveLM, AIDE, please contact and download from their source papers. The codes used in this study are available at DOI: 10.26599/ETSD.2026.9190024.

## Acknowledgements

The authors based in Quebec, Canada gratefully acknowledge financial support from the Fonds de recherche du Québec AUDACE International Program (FRQ AUDACE International), the Natural Sciences and Engineering Research Council of Canada (NSERC) Discovery Grant program, and the Canada Research Coordination Committee's New Frontiers in Research Fund (CRCC NFRF).

This research was also funded in part by the Luxembourg National Research Fund (FNR), grant reference: INR/AUDACE/25/19343746/EAMCIMCA. 

## Declaration of competing interest

The authors have no competing interests to declare that are relevant to the content of this article.

## Declaration of generative Al and AI-assisted technologies in the writing process

During the preparation of this work, the author(s) used ChatGPT-5.2 for language polishing. After utilizing this tool/service, the author(s) thoroughly reviewed and edited the content as necessary and take full responsibility for the final content of the published article.

## Author biography

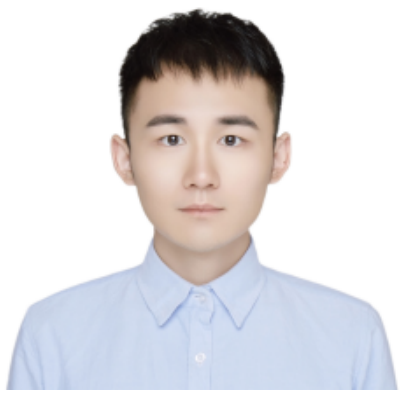

**Jiyao Wang** received the Bachelor degree from Sichuan University in 2021, M.Sc. degree from the Hong Kong University of Science and Technology (HKUST) in 2022, and a Ph.D. degree from HKUST, Guangzhou campus in 2025. Currently, he is a postdoctoral researcher at McGill University, Canada. His research interests include physiological signal measurement, intelligent transport systems, and human factors.

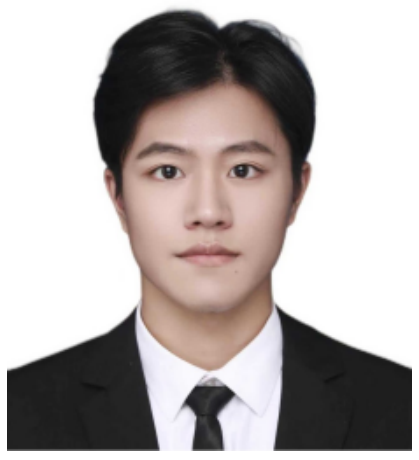

**Yunbiao Wang** received his Bachelor's and Master's degrees from Southwest Jiaotong University. He is currently a Ph.D. candidate at McGill University. His research focuses on intelligent transportation systems, including agentic mobility networks, human-AI interaction in transportation, and reinforcement learning for sustainable and resilient mobility solutions.

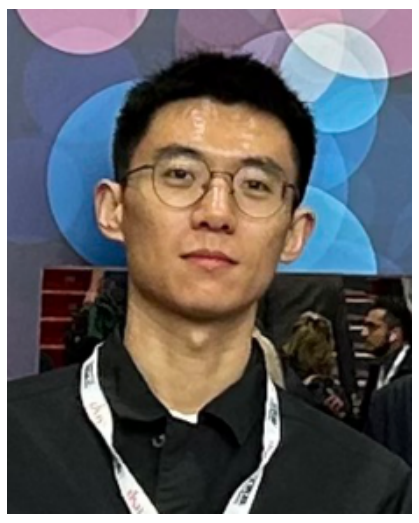

**Yubo Jiao** received his B.Sc. degree, M.Eng. degree, and Ph.D. degree from Southwest Jiaotong University, Chengdu, China, in 2016, 2019, and 2024, respectively. He is currently a postdoctoral researcher with the Department of Civil Engineering at McGill University, Canada. His research focuses on Human-Machine Collaborative Transportation, including driving behavior modeling, driver fatigue modeling, application of large language models in traffic planning and safety, data-driven traffic safety analysis.

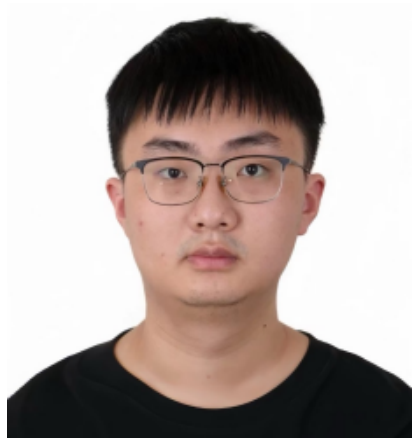

**Xiao Yang** is currently an M.Phil. student at the Hong Kong University of Science and Technology, Guangzhou campus. He received his bachelor's degree in Computing Science at Sichuan Agricultural University. His research interests include physiological signal measurement, state monitoring.

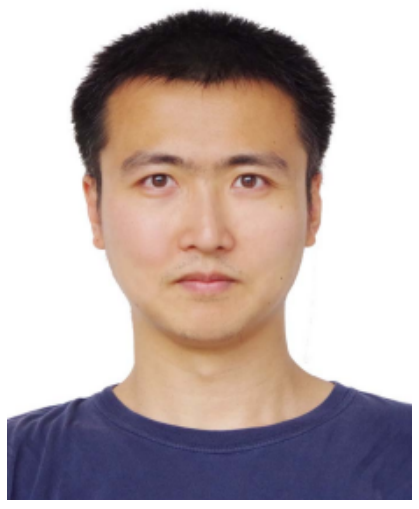

**Dengbo He** received his bachelor's degree from Hunan University in 2012, M.S. degree from Shanghai Jiao Tong University in 2016 and Ph.D. degree from the University of Toronto in 2020. He is currently an assistant professor at the Intelligent Transportation Trust and Robotics and Autonomous Systems Thrust, the HKUST(Guangzhou). He is also affiliated with the Department of Civil and Environmental Engineering, HKUST, Hong Kong SAR. From 2020 to 2021, he was a post-doctoral fellow at the University of Toronto.

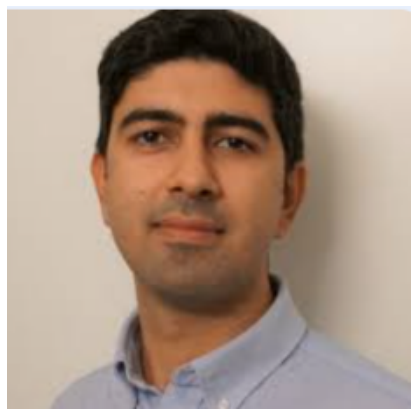

**Sasan Jafarnejad** obtained his PhD in Computer Science from the University of Luxembourg in 2020, where his doctoral research focused on modelling driver behaviour using various deep learning and machine learning algorithms. Following his PhD, he transitioned to industry, where he held a technical leadership role for three years. He subsequently returned to academia, joining the Ubiquitous and Intelligent Systems (UBIX) research group at the Interdisciplinary Centre for Security and Trust (SnT) at the University of Luxembourg as a Postdoctoral Researcher. His research interests include machine learning and artificial intelligence for transportation and mobility.

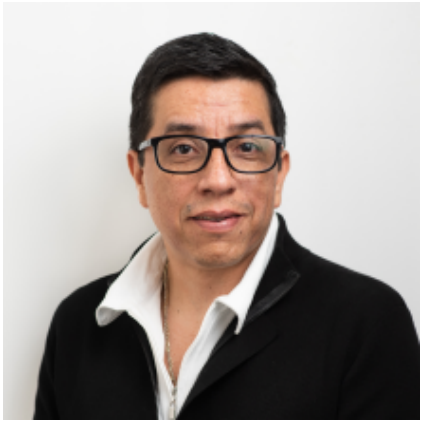

**Luis Miranda-Moreno** received the Ph.D. degree from the University of Waterloo, Waterloo, ON, Canada. He is currently an Associate Professor with the Department of Civil Engineering, McGill University, Montreal, QC, Canada. His research interests include the development of crash-risk analysis methods, the integration of emergency technologies for traffic monitoring, the impact of climate on transportation systems, the analysis of short- and long-term changes in travel demand, the impact of transport on the environment, evaluation of energy efficiency measures, and nonmotorized transportation.

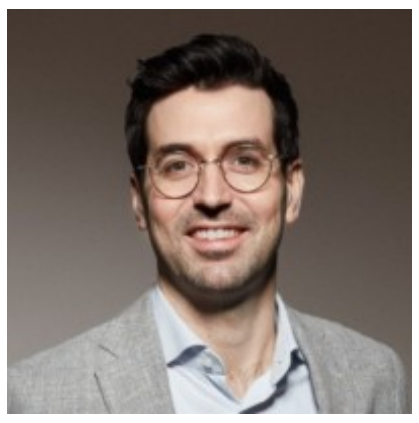

**Raphaël Frank** is an Associate Professor and Chief Research Scientist at the Interdisciplinary Centre for Security, Reliability and Trust (SnT), University of Luxembourg. He holds a Master's degree in Cryptography and Network Security from the University of Joseph Fourier in France, and a Ph.D. in Computer Science from the University of Luxembourg. During his Ph.D. studies, he was a visiting scholar at the University of California, Los Angeles (UCLA). He is the Head of the UBIX Research Group, which conducts research on distributed artificial intelligence and is the director of the IPBG ATLAS Industrial PhD Programme.

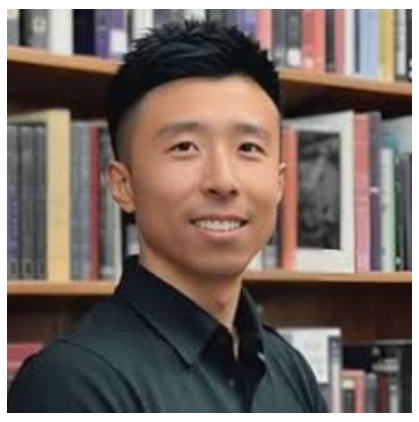

**Jiangbo Yu** is an Assistant Professor in the Department of Civil and Environmental Engineering at McGill University, Canada. He received the Ph.D. degree in Civil Engineering from the University of California, Irvine, the M.S. degree from the University of Southern California, and the B.Sc. degree from the Beijing Institute of Technology. Prior to joining McGill, he worked as a senior engineer at AECOM and Cambridge Systematics, and as a research associate at the Massachusetts Institute of Technology. His research focuses on the AI-enhanced planning and management of transportation system.